\RequirePackage{fix-cm}
\documentclass[twocolumn]{svjour3}          
\smartqed  
\usepackage{graphicx}
\usepackage{times}
\usepackage{epsfig}
\usepackage{amssymb}
\usepackage{multirow}
\usepackage{natbib}
\usepackage{algorithm}
\usepackage{algorithmic}
\usepackage{url}
\usepackage[svgnames]{xcolor}
\usepackage{marvosym}
\usepackage{amsmath}
\usepackage{hhline}

\newcommand{\norm}[1]{\left\lVert#1\right\rVert}

\usepackage[colorlinks]{hyperref}
\begin{document}

\title{Unsupervised learning of foreground object detection}


\author{Ioana Croitoru         \and
        Simion-Vlad Bogolin    \and
        Marius Leordeanu
}


\institute{
             Ioana Croitoru\textsuperscript{1} \at
              \email{ioana.croi@gmail.com}
            \and
            Simion-Vlad Bogolin\textsuperscript{1} \at
                \email{vladbogolin@gmail.com}
            \and 
            Marius Leordeanu\textsuperscript{1,2} \at
                \email{marius.leordeanu@imar.ro}
            \and 
                \textsuperscript{1}Institute of Mathematics of the Romanian Academy \at
                   \hspace*{1pt} 21 Calea Grivitei, Bucharest, Romania
            \and
              \textsuperscript{2}University "Politehnica" of Bucharest \at
                   \hspace*{1pt} 313 Splaiul Independentei, Bucharest, Romania
             \vspace*{-5pt}
            \and
               \noindent\rule{\linewidth}{0.4pt}
                   The first two authors contributed equally to this work.
}


\maketitle

\begin{abstract}
   Unsupervised learning poses one of the most difficult challenges in computer vision today. The task has an immense practical value with many applications in artificial intelligence and emerging technologies, as large quantities of unlabeled videos can be collected at relatively low cost. In this paper, we address the unsupervised learning problem in the context of detecting the main foreground objects in single images. We train a student deep network to predict the output of a teacher pathway that performs unsupervised object discovery in videos or large image collections. Our approach is different from published methods on unsupervised object discovery. We move the unsupervised learning phase during training time, then at test time we apply the standard feed-forward processing along the student pathway. This strategy has the benefit of allowing increased generalization possibilities during training, while remaining fast at testing. Our unsupervised learning algorithm can run over several generations of student-teacher training. Thus, a group of student networks trained in the first generation collectively create the teacher at the next generation. In experiments our method achieves top results on three current datasets for object discovery in video, unsupervised image segmentation and saliency detection. At test time the proposed system is fast, being one to two orders of magnitude faster than published unsupervised methods. 
   
   
\end{abstract}

\section{Introduction}\label{introduction}
  Unsupervised learning is one of the most difficult and interesting problems in computer vision and machine learning today. Many researchers believe that learning from large collections of unlabeled videos could help decode hard questions regarding the nature of intelligence and learning. Moreover, as unlabeled videos are easy to collect at relatively low cost, unsupervised learning could be of real practical value in many computer vision and robotics applications. In this article we propose a novel approach to unsupervised learning that successfully tackles many of the challenges associated with this task. We present a system that is composed of two main pathways, one that performs unsupervised object discovery in videos or large image collections along the teacher branch, and the other, the student branch, which learns from the teacher to detect foreground objects in single images. Our approach is general in the sense that the student or teacher pathways do not depend on a specific neural network architecture or implementation. Also, our approach allows the unsupervised learning process to continue over several generations of students and teachers. In Algorithm \ref{alg:general_unsup_learning} we present the high level description of our method. We will use throughout the paper the terms "generation" and "iteration" of Algorithm \ref{alg:general_unsup_learning} interchangeably. A preliminary version of this work, without presenting the possibility of learning over several generations and with fewer experimental results appeared at ICCV 2017 (\cite{croitoru2017unsupervised}).
  
  In Figure \ref{fig:system} we present a graphic overview of our full system. In the unsupervised training stage the student network (module A) learns, frame by frame, from an unsupervised teacher pathway (modules B and C) to produce similar object masks in single images. The student branch tries to imitate for each frame the output of the teacher, while having as input only a single image - the current frame. The teacher on the other hand has access to an entire video sequence. The method presented in Algorithm \ref{alg:general_unsup_learning} follows the main steps of the system as it learns from one iteration (generation) to the next.
  The steps are discussed in more detail in Section \ref{sec:approach}.
  
  During the first iteration of Algorithm \ref{alg:general_unsup_learning}, the unsupervised teacher pathway has access to information over time - a video. In contrast, the student is deeper in structure, but it has access only to a single image - the current video frame. Thus, the information discovered by the teacher in time is captured by the student in added depth, over neural layers of abstraction. Several student nets with different architectures are trained at the first iteration. In order to use as supervisory signal only good quality masks, an unsupervised mask selection procedure is applied, as explained in Section \ref{sec:systemarchitecture}. Once several student nets are trained, their output is combined to form the teacher at the next iteration. Then, we run, at the next generation, the newly formed teacher on a larger set of unlabeled videos, to produce supervisory signal for the next generation students.
  Note that while at the first iteration the teacher pathway is required to receive video sequences as input, from the second generation on, it could receive as input large image collections, as well. Due to the very high computational and storage costs, required during training time, we limit our experiments to learning over two generations, but our algorithm is general and could run over many iterations. We show in extensive experiments that even two generations are sufficient to significantly outperform the current state of the art on object discovery in video and images. We also demonstrate a solid improvement from one generation to the next. Now we enumerate the main contributions of our approach: 
  
\begin{figure}[t]
\begin{center}
   \includegraphics[width=1\linewidth]{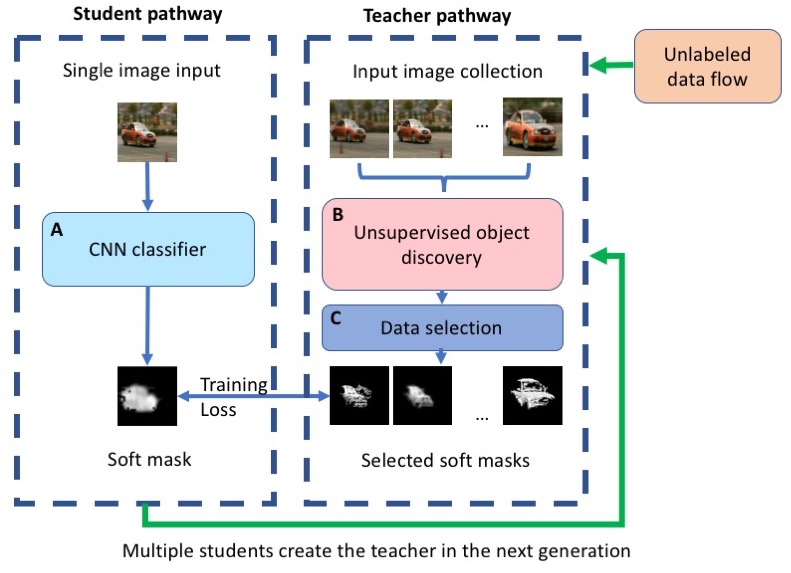}
\end{center}
   \caption{The dual student-teacher system proposed for unsupervised learning to detect foreground objects in images, functioning as presented in Algorithm \ref{alg:general_unsup_learning}. It has two pathways: along the teacher branch, an object discoverer in videos or large image collections (module B) detects foreground objects. The resulting soft masks are then filtered based on an unsupervised data selection procedure (module C). The resulting final set of pairs - input image (a video frame) and soft mask for that particular frame (which acts as an unsupervised label) - are used to train the student pathway (module A). The whole process can be repeated over several generations. At each generation several student CNNs are trained, then they collectively contribute to form a more powerful teacher, at the next iteration of the overall algorithm. }
\label{fig:system}
\end{figure}

\textbf{1)} We introduce a novel approach to unsupervised learning from videos to detect foreground objects in images. The overview of our system and algorithm are presented in Figure \ref{fig:system} and Algorithm \ref{alg:general_unsup_learning}. The system has two main pathways - one that acts as a teacher and discovers objects in videos or large collections of images and the other that acts as student and learns from the teacher to detect the foreground objects in single input images. We provide a general algorithm for unsupervised learning over several generations of students and teachers. We experiment with different types of student nets and show how they collectively work together to form the teacher at the next generation. This is done in conjunction with a novel unsupervised soft-mask selection scheme. We demonstrate experimentally that within a generation the students are more powerful than their teachers, while both pathways improve significantly from one generation to the next.

\textbf{2)} At the higher level, our proposed algorithm is sufficiently general to accommodate different implementations and neural network architectures. In this paper, we also provide a specific implementation which we describe in detail. We demonstrate its performance on three recent datasets, namely YouTube Objects (\cite{prest2012learning}), Object Discovery in Internet Images (\cite{rubinstein2013unsupervised}) and Pascal-S (\cite{li2014secrets}), on which we obtain state of the art results. To our best knowledge, it is the first system that learns to detect and segment foreground objects in images in unsupervised fashion, with no pre-trained features given or manual labeling, while requiring only a single image at test time.
  
\section{Scientific context}
\label{stoa}
The literature on unsupervised learning follows two main directions. 1) One is to learn powerful features in an unsupervised way and then use them for transfer learning, within a supervised scheme and in combination with different classifiers, such as SVMs or CNNs (\cite{radenovic2016cnn, misra2016shuffle, li2016unsupervised}). 
2) The second direction is to discover, at test time, common patterns in unlabeled data, using clustering, feature matching or data mining formulations (\cite{jain1999data,cho2015unsupervised,key:sivic_05}).

Belonging to the first category and closely related to our work, the approach in~\cite{pathak2017learning} proposes a system in which a deep neural network learns to produce soft object masks from an unsupervised module that uses optical flow cues in video. The deep features learned in this manner are then applied to several transfer learning tasks. Different from their work, we provide a more general approach that could learn in an unsupervised manner over several generations. 
From an experimental point of view, while~\cite{pathak2017learning} tests their work on a supervised transfer learning task, we evaluate ours on specific unsupervised foreground object detection and segmentation tasks and demonstrate state of the art performance, often by a large margin.

Recently, researchers have started to use the natural, spatial and temporal structure in images and videos as supervisory signals in unsupervised learning approaches that are considered to follow a~\textit{self-supervised learning} paradigm (\cite{raina2007self, lee2017unsupervised, wang2015unsupervised}). Methods that fall into this category include those that learn to estimate the relative patch positions in images (\cite{doersch2015unsupervised}), predict color channels (\cite{larsson2016learning}), 
solve jigsaw puzzles (\cite{noroozi2016unsupervised}) and inpaint (\cite{pathak2016context}). One trend is to use as supervisory signal, spatial and appearance information collected from raw single images. In such single-image cases the amount of information that can be learned is limited to a single moment in time, as opposed to the case of learning from video sequences. Using unlabeled videos as input is closer related to our work and includes learning to predict the temporal order of frames (\cite{lee2017unsupervised}), generate the future frame (\cite{finn2016unsupervised, xue2016visual, goroshin2015learning}) or learn from optical flow (\cite{Wang_2015_ICCV}).

For most of these papers, the unsupervised learning scheme is only an intermediate step to train features that are eventually used on classic supervised learning tasks, such as object classification, object detection or action recognition. Such pre-trained features perform better than randomly initialized ones, as they contain valuable semantic information implicit in the natural structure of the world used as supervisory signal. In our work, we focus mostly on specific unsupervised tasks on which we perform extensive evaluations, but we also show some results on transfer learning experiments.

The second main approach to unsupervised learning includes methods for image co-segmentation  (\cite{joulin2010discriminative,kim2011distributed,rubinstein2013unsupervised,joulin2012multi,kuettel2012segmentation,vicente2011object,rubio2013video,leordeanu2012unsupervised}) and weakly supervised localization (\cite{deselaers2012weakly,nguyen2009weakly,siva2013looking}). Earlier methods are based on local feature matching and detection of their co-occurrence patterns (\cite{stretcu2015multiple,key:sivic_05,key:leordeanu_cvpr05,key:parikh_07_2,liu2007topic}), while more recent ones (\cite{joulin2014efficient,rochan2014efficient}) discover object tubes by linking candidate bounding boxes between frames with or without refining their location. Traditionally, the task of unsupervised learning from image sequences has been formulated as a feature matching or data clustering optimization problem, which is computationally very expensive due to its combinatorial nature. 

There are also other 
papers (\cite{lee2011key,Cheng_2017_ICCV,Jain_2017_CVPR,Tokmakov_2017_CVPR}) that tackle unsupervised learning tasks but are not fully unsupervised, using powerful features that are pre-trained in supervised fashion on large datasets, such as ImageNet (\cite{russakovsky2015imagenet}) or VOC2012 (\cite{Everingham15}). Such works take advantage of the rich source of supervised information learned from other datasets, through features trained to respond to general object properties over tens or hundreds of object categories. 

With respect to the end goal, our work is more related to the second research direction, on unsupervised discovery in video. However, unlike that research, we do not discover objects at test time, but during the unsupervised training process, when the student pathway learns to detect foreground objects. Therefore, from the learning perspective, our work is more related to the first research direction based on self-supervised training.

\section{Overall approach}
  \label{sec:approach}
  
  We propose a genuine unsupervised learning algorithm for foreground object detection that offers the possibility to improve over several iterations. Our method combines in complementary ways multiple modules that are well suited for this task. It starts with a teacher pathway that discovers objects in unlabeled videos and produces a soft mask of the foreground object in each frame. The resulting soft-masks of lower quality are then filtered out automatically. Next, the remaining ones are passed to a student ConvNet, which learns to predict object masks in single images. When several student nets of different architectures are learned they form a new teacher for the next generation, then the whole process is repeated. At the next iteration we bring in more unlabeled data, we learn in an unsupervised fashion a better data selection mechanism and ultimately train more powerful student networks. In Algorithm \ref{alg:general_unsup_learning} we enumerate concisely the main steps of our approach. 
  
 \begin{algorithm}[t!]
  \caption{Unsupervised learning of foreground object detection}
        \label{alg:general_unsup_learning}
        \begin{algorithmic}
      \STATE \textbf{Step 1:} perform unsupervised object discovery in unlabeled videos, along the teacher pathway (module B in Figure \ref{fig:system}).
      \STATE \textbf{Step 2:} automatically filter out poor soft masks produced at the previous step (module C in Figure \ref{fig:system}).
      \STATE \textbf{Step 3:} use the remaining masks as supervisory signal for training one or more student nets, along the student pathway 
      (module A in Figure \ref{fig:system}).
      \STATE \textbf{Step 4:} use the ensemble of student nets to form a new teacher and learn a more powerful soft-mask selector, for the next iteration (referred to as a novel student-teacher generation).
      \STATE \textbf{Step 5:} extend the unlabeled video dataset and return to Step 1 to train the next generation (note that from this step forward, the training dataset can also be extended with collections of unlabeled images, not just videos). 
  \end{algorithmic}
  \end{algorithm}

  Now we present the main algorithm in more detail. At Step 1 we start with an object discoverer in video sequences. There are several available methods for video discovery in the literature, with good performance (\cite{borji2012salient, cheng2015global, barnich2011vibe}). We chose the VideoPCA algorithm introduced as part of the system in~\cite{stretcu2015multiple} because it is very fast (50-100 fps), uses very simple features (individual pixel colors) and it is completely unsupervised, with no usage of supervised pre-trained features. It learns how to separate the foreground from the background. It exploits the spatio-temporal consistency in appearance, shape, movement and location of objects, common in video shots, along with the 
  contrasting properties, in size, shape, motion and location, between the main object and the background scene. Note that it would be much harder, at this first stage, to discover objects in collections of unrelated images, where there is no smooth variation in shape, appearance and location over time. Only at the second iteration of the algorithm, the simpler VideoPCA is replaced with a more powerful ensemble of student nets which is able to discover objects in collections of images as well. 
  
  The teacher branch produces soft foreground masks, one per each frame, which are not always of good quality. Thus, at Step 2, we use, during the first iteration, a simple and effective way to filter out poor masks. Only at the second iteration we are able to learn a more powerful soft-mask selector (see Section \ref{sec:data_selection}). The soft-masks that pass the filtering phase are then used (Algorithm \ref{alg:general_unsup_learning}, Step 3) to train the student pathway. As we want the student branch to learn general visual properties of objects in images, we limit its access to a single input image. 
  
  Our approach offers the possibility of improving performance by training a next generation of object detectors. In experiments, we found that there are three key aspects, which are effective at improving generalization at the next iteration: 1) we need to train several student nets (at module A), preferably of different architectures, which are stronger in combination than separately. Then, they become the teacher (module B) at the next iteration; 2) we train, also in an unsupervised fashion, a better soft-mask selector (module C); 3) it is preferred to increase the unlabeled training set at the next iteration, for improved generalization.
  
  Having access to the complete training set at the very first iteration could be useful, but it is not optimal. At that stage, the teacher is still weak and imposes a certain limitation on how much could be learned from the data, no matter how large that data is. Getting access to a larger unlabeled training dataset is more effective at the second iteration, when the teacher pathway is significantly stronger. The idea of gradually increasing the complexity in the training set is also related to curriculum learning (\cite{bengio2009curriculum}), when we start with simpler cases then add more difficult ones. Increasing the strength of the teacher pathway improves the quality of the supervisory signal, while introducing more unlabeled data increases variety. Both act together in order to improve generalization.
\section{System architecture}\label{sec:systemarchitecture}

We, now, detail the architecture and training process of our system, module by module, as seen in Figure \ref{fig:system}. We first present the student pathway (module A in Figure \ref{fig:system}), which takes as input an individual image (e.g. current frame in the video) and learns to predict foreground soft-masks from an unsupervised teacher. The teacher pathway (represented by modules B and C in Figure \ref{fig:system}), is explained in detail in the Section \ref{sec:teacher}. 

\begin{figure*}
\begin{center}
   \includegraphics[width=1.0\linewidth]{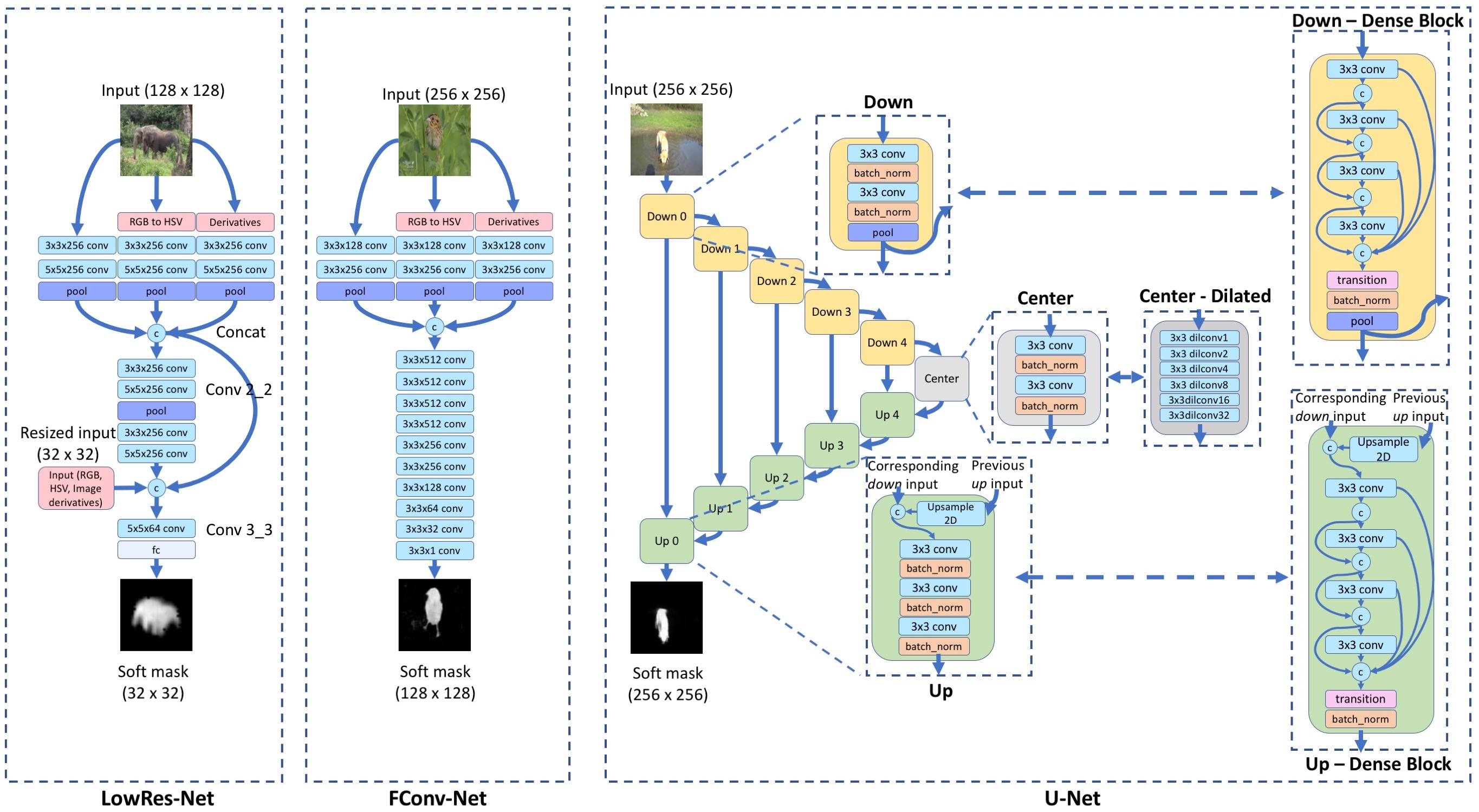}
\end{center}
   \caption{Different architectures for the "student" networks, each processing a single image. They are trained to predict the unsupervised label masks given by the teacher pathway, frame by frame. The architectures vary from the more classical baseline LowRes-Net (left), with low resolution output, to more recent architectures, such as the fully convolutional one (middle) and different types of U-Nets (right).  For the U-Net architecture the blocks denoted with double arrows can be interchanged to obtain a new architecture. We noticed that on the task of bounding box fitting the simpler low-resolution network performed very well, while being outperformed by the U-Nets on fine object segmentation.}
\label{fig:network_all}
\end{figure*}

\vspace*{-5pt}

\subsection{Student path: single-image segmentation}
\label{sec:cnn}

The student processing pathway (module A in Figure \ref{fig:system}) consists
of a deep convolutional network. We test different neural network architectures, some of which are commonly used in the recent literature on semantic image segmentation. We create a small pool of relatively diverse architectures, presented next.

The first convolutional network architecture for semantic segmentation that we test, is based on a more traditional CNN design. We term it LowRes-Net (see Figure~\ref{fig:network_all}) due to its low resolution soft-mask output. It has ten layers (seven convolutional, two pooling and one fully connected) and skip connections. Skip connections have proved to offer a boost in performance, as shown in the literature (\cite{raiko2012deep, pinheiro2016learning}). We also observed a similar improvement in our experiments when using skip connections. The LowRes-Net takes as input a $128\times128$ RGB image (along with its hue, saturation and derivatives w.r.t. x and y) and produces a $32\times32$ soft segmentation of the main objects present in the image.  Because LowRes-Net has a fully connected layer at the top, we reduced the output resolution of the soft-segmentation mask, to limit memory cost. While the derviatives w.r.t x and y are in principle not needed (as they could be learned by appropriate filters during training), in our tests explicitly providing the derivatives along with HSV and by using skip-connections boosted the accuracy by over $1\%$. The LowRes-Net has a total of 78M parameters, most of them being in the last, fully connected layer.

The second CNN architecture tested, termed FConv-Net, is fully convolutional (\cite{long2015fully}), as also presented in Figure \ref{fig:network_all}.
It has a higher resolution output of 128x128, with input size 256x256. Its main structure is derived from the basic LowRes-Net model. Different from LowRes-Net, it is missing the fully connected layer at the end and has more parameters in the convolutional layers, for a total of 13M parameters.

We also tested three different nets based on the U-Net (\cite{ronneberger2015u}) architecture, which proved very effective in the semantic segmentation literature. Our U-net networks are: 1) BasicU-Net, 2) DilateU-Net - similar to BasicU-Net but using atrous (dilated) convolutions (\cite{yu2015multi}) in the \textit{center} module, and 3) DenseU-Net - with dense connections in the \textit{down} and \textit{up} modules (\cite{jegou2017one}). 

The BasicU-Net has 5 \textit{down} modules with 2 convolutional layers each, with 32, 64, 128, 256 and 512 features maps, respectively. In the \textit{center} module the BasicU-Net has two convolutional layers with 1024 feature maps each. The \textit{up} modules have 3 convolutional layers and the same number of features maps as the corresponding \textit{down} modules. The only difference between BasicU-Net and DilateU-Net is that the former has a different \textit{center} module with 6 atrous convolutions and 512 feature maps each. Then, DenseU-Net has 4 \textit{down} modules with 4 corresponding \textit{up} modules. Each \textit{down} and \textit{up} module has 4 convolutions with skip-connections (as presented in Figure \ref{fig:network_all}). The modules have 12, 24, 48 and 64 features maps, respectively. The \textit{transition} represents a convolution, having the role of reducing the output number of feature maps from each module. The BasicU-Net has 34M parameters, while the DilateU-Net has 18M parameters. DenseU-Net has only 3M parameters, but uses skip-connections inside the \textit{up} and \textit{down} blocks in order to make up for the difference in the number of parameters. 
All three U-Nets have 256x256 input and same resolution output. 
All networks use ReLU activation functions. Please see Figure \ref{fig:network_all} for more specific details regarding the architectures of the different models. 

Given the current setup, the student nets do not learn to identify specific object classes. They will learn to softly segment the main foreground objects present, regardless of their particular category. The main difference in their performance is in their ability to produce fine object segmentations. While the LowRes-Net tends to provide a good support for estimating the object's bounding box due to its simpler output, the other ConvNets (especially the U-Nets), with higher resolution, are better at finely segmenting objects. Due to the different ways in which the particular models make mistakes, they are always stronger when forming an ensemble. In experiments we also show that they outperform their teacher and are able to detect objects from categories that were not seen during training.

\subsubsection{Student networks ensemble}
\label{sec:ensemble}

The pool of student networks with different architectures produce varied results that differ qualitatively. While the bounding boxes computed from their soft-masks have similar accuracy, the actual soft-segmentation output looks differently. They have different strengths, while making different kinds of mistakes. The above observation immediately suggests that they should be stronger in combination, so we have experimented with the idea of combining them into an ensemble. We propose two types of ensembles.

The first one, termed Multi-Net, outputs a soft-mask that is obtained by multiplying pixel-wise the soft-masks produced by each individual student net. Thus, only positive pixels, on which all nets agree, survive to the final segmentation. Multi-Net offers robust masks of significantly higher quality. In Section \ref{sec:data_selection} we show how Multi-Net can be effectively used to learn in an unsupervised fashion, a network (EvalSeg-Net) for evaluating the goodness of a specific segmentation. That network is an important part of the next generation teacher pathway and replaces module C at the next iteration.

The second approach to forming an ensemble is to use EvalSeg-Net in order to select the best soft-mask from the pool of masks generated by the student nets. We term this ensemble system, MultiSelect-Net. Quantitatively, MultiSelect-Net and Multi-Net perform similarly, but Multi-Net tends to produce fuzzier masks due to the additional multiplication of the student's soft-masks. 

\subsubsection{Training the student ConvNets}
\label{sec:training_student_nets}

We treat foreground object segmentation as a multidimensional regression problem,
where the soft mask given by the unsupervised video segmentation system acts as the
desired output. Let $\mathbf{I}$ be the input RGB image (a video frame) and $\mathbf{Y}$ be the
corresponding 0-255 valued soft segmentation given by the unsupervised teacher for that particular frame. 
The goal of our network is to predict a soft segmentation mask
$\mathbf{\hat{Y}}$ of width $W$ and height $H$ (where $W=H=32$ for the basic architecture, $W=H=128$ for fully convolutional architecture and $W=H=256$ for U-Net architectures), that approximates as well as possible the mask 
$\mathbf{Y}$. For each pixel in the output image, 
we predict a 0-255 value, so that the total difference between 
$\mathbf{Y}$ and $\mathbf{\hat{Y}}$ is minimized. Thus, given a set of 
$N$ training examples, let $\mathbf{I}^{(n)}$ be the input image (a video frame), ${\mathbf{\hat{Y}}}^{(n)}$ be the 
predicted output mask for $\mathbf{I}^{(n)}$, 
$\mathbf{Y}^{(n)}$ the soft segmentation mask (corresponding to $\mathbf{I}^{(n)}$) and  $\mathbf{w}$ the network parameters.  $\mathbf{Y}^{(n)}$ is produced by the video discoverer after processing the video that $\mathbf{I}^{(n)}$ belongs to. Then, our loss is:

\vspace*{-10pt}
\begin{equation}
\label{eq:learning}
L(\mathbf{w})=\frac{1}{N}\sum\limits_{n=1}^N \sum\limits_{p=1}^{W\times H}{(\mathbf{Y}_{p}^{(n)} - \mathbf{\hat{Y}}_{p}^{(n)}(\mathbf{w}, \mathbf{I}^{(n)}))}^2
\end{equation}

where $\mathbf{Y}_{p}^{(n)}$ and $\mathbf{\hat{Y}}_{p}^{(n)}$ denotes the $p$-th pixel from $\mathbf{Y}^{(n)}$, respectively $\mathbf{\hat{Y}}^{(n)}$. 

We observed that in our tests, the L2 loss performed better than the cross-entropy loss, due to the fact that the soft-masks used as labels have real values, not discrete ones. Also, they are not perfect, so the idea of thresholding them for training does not perform as well as directly predicting their real values.
We train our network using the Tensorflow (\cite{abaditensorflow}) framework with the Adam optimizer (\cite{kingma2014adam}). All models are trained end-to-end using a fixed learning rate of 0.001 for 10 epochs. The training time for any given model is about 3-5 days on a Nvidia GeForce GTX 1080 GPU, for the first iteration and about 2 weeks for the second iteration students. \\

\noindent \textbf{Post-processing.} The student CNN outputs a $W \times H$ soft mask. In order to fairly compare our models with other methods, we have two different post processing steps: 1) bounding box fitting and 2) segmentation refinement. For fitting a box around the soft mask, we first up-sample the $W \times H$ output to the original size of the image, then threshold the mask (validated on a small subset), determine the connected components and fit a tight box around each of the components. We perform segmentation refinement (point 2) in a single case, on the Internet Images Dataset as also specified in the experiments section. For that, we use
the OpenCV implementation of GrabCut (\cite{rother2004grabcut}) to refine our soft mask, up-sampled to the original size. In all other tests we use the original output of the networks.

\subsection{Teacher path: unsupervised discovery in video}
\label{sec:teacher}

There are several methods
available for discovering objects and salient regions in images and videos (\cite{borji2012salient,cheng2015global,hou2007saliency, jiang2013salient,cucchiara2003detecting,barnich2011vibe})
with reasonably good performance. More recent methods for foreground objects discovery such as \cite{papazoglou2013fast} are both relatively fast and accurate, with runtime around $4$ seconds per frame. However, that runtime is still long and prohibitive for training the student CNN that requires millions of images. For that reason we used at the first generation (Iteration 1 of Algorithm \ref{alg:general_unsup_learning}) for module B in Figure \ref{fig:system}, the VideoPCA algorithm, which is a part of the whole system introduced in \cite{stretcu2015multiple}. It has lower accuracy than the full system, but it is much faster, running at $50-100$ fps. At this speed we can produce one million unsupervised soft segmentations in a reasonable time of about 5-6 hours. \\

\noindent \textbf{VideoPCA.} The main idea behind VideoPCA is to model the background in video frames with Principal Component Analysis. It finds initial foreground regions as parts of the frames that are not reconstructed well with the PCA model. Foreground objects are smaller than the background,
have contrasting appearance and more complex movements. They could be seen as outliers, within the larger background scene. That makes them less likely
to be captured well by the first PCA components. Thus, for each frame, an initial soft-mask is produced from an error image, which is the difference between the original image and the PCA reconstruction. These error images are first smoothed with a large Gaussian filter and then thresholded. The binary masks obtained are used to learn color models of foreground and background, based on which individual pixels are classified as belonging to foreground or not. The object masks obtained are further multiplied with a large centered Gaussian, based on the assumption that foreground objects are often closer to the image center. These are the final masks used in your system. For more technical details, the reader is invited to consult \cite{stretcu2015multiple}. In this work, we use the method exactly as found online\footnote{\url{https://sites.google.com/site/multipleframesmatching/}} without any parameter tuning. \\

\noindent \textbf{Teacher pathway at the next generation:}
At the next iteration of Algorithm \ref{alg:general_unsup_learning}, 
VideoPCA (in module B) is replaced by the student nets trained at the previous iteration in the following way. While we could use as new module B any of the two ensembles Multi-Net or MultiSelect-Net, we preferred a simpler and more efficient approach. For each unlabeled training image we ran all student nets and obtain multiple soft-masks, without combining them to produce a single output per image. 
Therefore the new module B is the collection of all student nets acting in parallel. Then, their soft-masks are filtered independently (using a given threshold) by the new Module C in Figure \ref{fig:system}, which is represented at the second iteration by EvalSeg-Net. Note that it is possible in this manner to obtain one, several or no soft segmentations for a given training image.
This approach is fast and it offers the advantage of processing data in parallel over multiple GPUs, without having to wait for all student nets to finish for every input image. As our experiments demonstrate, the approach is also efficient, with 
significantly better results at the second generation.

\subsubsection{Unsupervised soft masks selection}
\label{sec:data_selection}

\begin{figure*}[t]
\begin{center}
   \includegraphics[width=\linewidth]{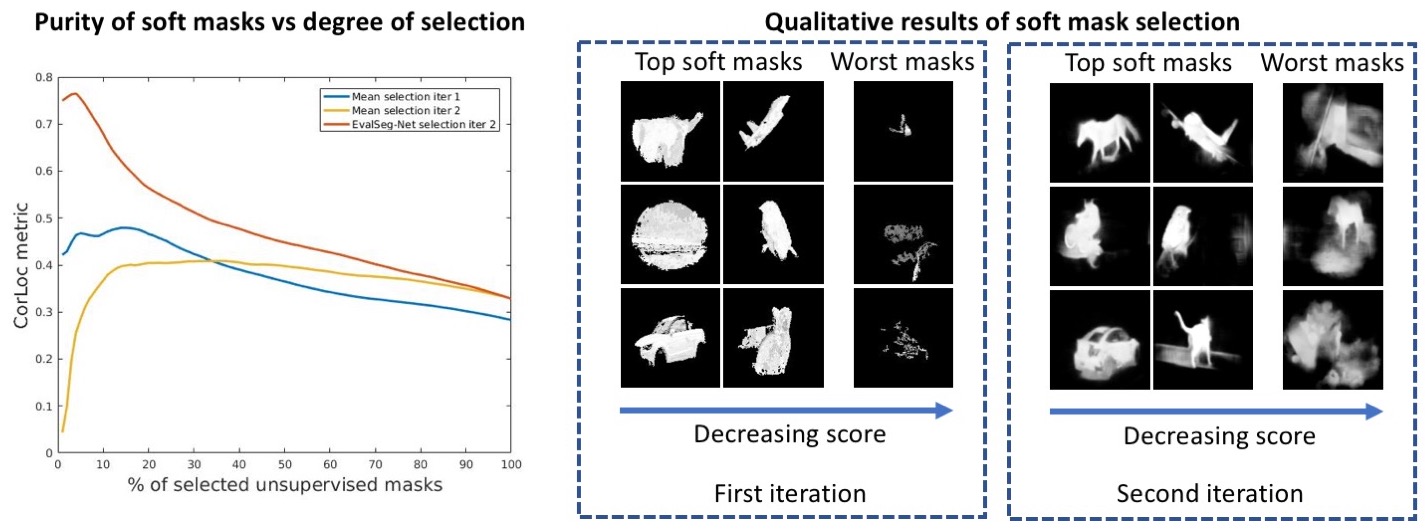}
\end{center}
   \caption{Purity of soft masks vs. degree of selection. When selectivity increases, the true purity of the training frames improves. Our automatic selection method is not perfect: some low quality masks may have high scores, while other good ones may be ranked lower. At the first iteration of Algorithm \ref{alg:general_unsup_learning} we select masks obtained with VideoPCA, while at the second generation we selected masks obtained with the teacher at the second generation. The plots are computed using results from the VID dataset, where there is an annotation for each input frame. Note the significantly better quality of masks at the second iteration (red vs. blue lines, in the left plot). We have also compared the simple "mean" based selection procedure used at iteration 1 (yellow line) with EvalSeg-Net used at iteration 2 (red line), on the same soft masks from iteration 2. The EvalSeg-Net is more powerful, which justifies its use at the second iteration when it replaces the simple "mean" based procedure.}
\label{fig:data_augmentation}
\end{figure*}

\begin{figure}[t]
\begin{center}
   \includegraphics[width=\linewidth]{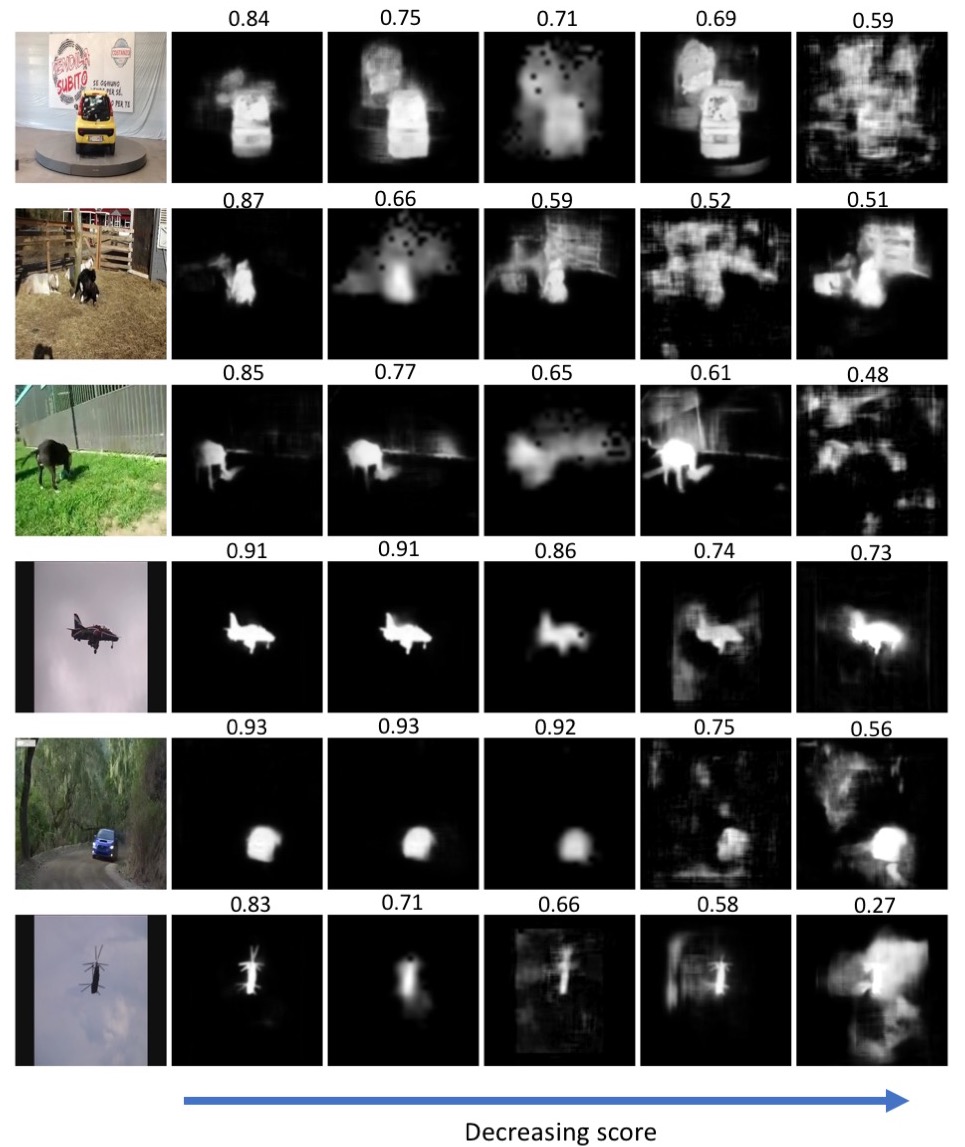}
\end{center}
   \caption{Qualitative results of the unsupervised EvalSeg-Net used for measuring segmentation "goodness" and filtering bad masks (Module C, iteration 2). For each input image we present five soft-masks candidates (from first iteration students) along with their "goodness" scores given by EvalSeg-Net, in decreasing order of scores. Note the effectiveness of EvalSeg-Net at ranking soft segmentations.}
\label{fig:confidence_visual}
\end{figure}

The performance of the student net is influenced by the quality of the
soft masks provided as labels by the teacher branch. 
The cleaner the masks, the more chances the student has to learn to segment well objects in images. VideoPCA tends to produce good results if the object present in the video stands out well against the background scene, in terms of motion and appearance. However, if the object
is occluded at some point, does not move w.r.t the scene or has a similar appearance to its background, the resulting soft masks might be poor. In the first generation, we used a simple measure of masks quality to select only the good soft-masks for training the student pathway, based on the following observation: when VideoPCA masks are close to the ground truth, the average of their nonzero values is usually high. Thus, when the discoverer is confident, 
it is more likely to be right. The average value of non-zero pixels in the soft mask 
is then used as a score indicator for each segmented frame. Only masks of certain quality according to this indicator are selected and used for training the student nets.
This represents module C in Figure \ref{fig:system} at the first generation of Algorithm \ref{alg:general_unsup_learning}. 
While being effective at iteration 1, the simple average value over all pixels cannot capture the goodness of a segmentation at the higher level of overall shape. At the next iterations, we therefore explore new ways to improve it.

Consequently, at the next iterations we propose an unsupervised way for learning the EvalSeg-Net to estimate segmentation quality. As mentioned previously, Multi-Net provides masks of higher quality as it cancels errors from individual student nets. Thus, we use the cosine similarity between a given individual segmentation and the ensemble Multi-Net mask, as a cost for "goodness" of segmentation. Having this unsupervised segmentation cost we
train the EvalSeg-Net deep neural net to predict it. As previously mentioned, this net acts as an automatic mask evaluation procedure, which in subsequent iterations becomes module C in Figure \ref{fig:system}, replacing the simple mask average value used at Iteration 1. Only masks that pass a certain threshold are used for training the student path.

The architecture of EvalSeg-Net is similar to LowRes-Net (Figure \ref{fig:network_all}), with the difference that the input channel containing image derivatives is replaced by the actual soft-segmentation that requires evaluation and it does not have skip connections. Also, after the last fully connected layer (size 512) we add a last one-neuron layer to predict the segmentation quality score, which is a single real valued number. 

Let $\mathbf{I}$ be an input RGB image, $\mathbf{S}$ an input soft-mask, $\mathbf{\hat{Y}}=\prod_{i=1}^{5}{\mathbf{\hat{Y}}_{N_i}}$ be the output of our Multi-Net where $\mathbf{\hat{Y}}_{N_i}$ denotes the output of network $N_i$. We treat the segmentation "goodness" evaluation task as a regression problem where we want to predict the Cosine similarity between $\mathbf{S}$ and $\mathbf{\hat{Y}}$. So, our loss for EvalSeg-Net is defined as follows:

\begin{equation}
\label{eq:multi-seg-loss}
L(\mathbf{w})=\frac{1}{K}\sum\limits_{k=1}^K {\biggl(\hat{o}^{(k)}(\mathbf{w}, \mathbf{I}^{(k)}, \mathbf{S}^{(k)}) - 
\frac{\mathbf{S}^{(k)} \cdotp \mathbf{\hat{Y}}^{(k)}}{\norm{\mathbf{S}^{(k)}} \norm{\mathbf{\hat{Y}}^{(k)}}} \biggr)}^2
\end{equation}

where $K$ represents the number of training examples and $\hat{o}^{(k)}(\mathbf{w}, \mathbf{I}^{(k)}, \mathbf{S}^{(k)})$ represents the output of EvalSeg-Net for image $\mathbf{I}^{(k)}$ and soft mask $\mathbf{S}^{(k)}$. 

Given a certain metric for segmentation evaluation (depending on the learning iteration), we keep only the soft masks above a threshold for each dataset (e.g. VID (\cite{russakovsky2015imagenet}), YTO (\cite{prest2012learning}), Youtube Bounding Boxes (\cite{real2017youtube})). In the first iteration this threshold was obtained by sorting the VideoPCA soft-masks based on their score and keeping only the top 10 percentile, while on the second iteration we validate a threshold ($=0.8$) on a small dataset and select each mask independently by using this threshold on the single value output of EvalSeg-Net. \\

\noindent \textbf{Mask selection evaluation.}
In Figure \ref{fig:data_augmentation} we present the dependency of segmentation performance w.r.t ground truth object boxes (used only for evaluation) vs. the percentile $p$ of masks kept after the automatic selection, for both generations. We notice the strong correlation between the percentage of frames kept and the quality of segmentations. It is also evident that the EValSeg-Net is vastly superior to the simpler procedure used at iteration 1. EvaSeg-Net is able to correctly evaluate soft segmentations even in more complex cases (see Figure \ref{fig:confidence_visual}).

Even though, we can expect to improve the quality of the unsupervised masks by drastically pruning them (e.g. keeping a smaller percentage), the fewer we are left with, the less training data we get, increasing the chance to overfit. We make up for the losses in training data by augmenting the set of training masks and by also enlarging the actual unlabeled training set at the second generation. There is a trade-off between level of selectivity and training data size: the more selective we are about what masks we accept for training, the more videos we need to collect and process through the teacher pathway, to obtain the sufficient training data size. \\

\noindent \textbf{Data augmentation.}
A drawback of the teacher at the first learning iteration (VideoPCA) is that it can only detect the main object if it is close to the center of the image. The assumption that the foreground is close to the center is often true and indeed helps that method, which has no deep learned knowledge, 
to produce soft masks with a relatively high precision. Not surprisingly, it often fails when the object is not in the center, therefore its recall is relatively low.
Our data augmentation procedure addresses this limitation and can be concisely described as follows: randomly crop patches of the input image, 
covering 80\% of the original image and scale up the patch to the expected input size.
This produces slightly larger objects at locations that cover the whole image area, not just the center. As experiments show, the student net is able to see objects at different locations in the image, unlike its raw teacher (VideoPCA at iteration 1), which is strongly biased towards the image center.

At the second generation, the teacher branch is significantly better at detecting objects at various locations and scales in the image. Therefore, while artificial data augmentation remains useful (as it is usually the case in deep learning), its importance diminishes at the second iteration of learning (Algorithm \ref{alg:general_unsup_learning}).

\subsection{Implementation pipeline}
\label{sec:pipeline}

Now that we have presented in technical detail all major components of our system, we concisely present the actual steps taken in our experiments, in sequential order, and show how they relate to our general Algorithm \ref{alg:general_unsup_learning} for unsupervised learning to detect foreground objects in images. 

\begin{enumerate}
    \item Run VideoPCA on input images from VID and YouTube Objects datasets (Algorithm \ref{alg:general_unsup_learning}, Iteration 1, Step 1)
    \item Select VideoPCA masks using first generation selection procedure (Algorithm \ref{alg:general_unsup_learning}, Iteration 1, Step 2)
    \item Train first generation student ConvNets on the selected masks, namely LowRes-Net, FConv-Net, BasicU-Net, DilateU-Net and DenseU-Net (Algorithm \ref{alg:general_unsup_learning}, Iteration 1, Step 3).
    \item Create first generation student ensemble Multi-Net by multiplying the outputs of all students and train
    EvalSeg-Net to predict the similarity between a particular mask and the mask of Multi-Net. Create the second ensemble MultiSelect-Net by using EvalSeg-Net in combination with the student's masks (Algorithm \ref{alg:general_unsup_learning}, Iteration 1, Step 4).
    \item Add new data from YouTube Bounding Boxes.  
    (Algorithm \ref{alg:general_unsup_learning}, Iteration 1, Step 5)
    \item Return to Step 1, the teacher pathway: predict multiple soft-masks per input image on the enlarged unlabeled video set, using the student nets from Iteration 1 (Module B, Iteration 2), which will be then selected with EvalSeg-Net at Module C.
    (Algorithm \ref{alg:general_unsup_learning}, Iteration 2, Step 1)
    \item Select only sufficiently good masks evaluated with EvalSeg-Net (Algorithm \ref{alg:general_unsup_learning}, Iteration 2, Step 2)
    \item Train the second generation students on the newly selected masks. We use the same architectures as in Iteration 1 (Algorithm \ref{alg:general_unsup_learning}, Iteration 2, Step 3)
    \item Create the second generation student ensembles Multi-Net and MultiSelect-Net.
    (Algorithm \ref{alg:general_unsup_learning}, Iteration 2, Step 4)
\end{enumerate}

\vspace*{3pt}
The method presented in the introduction sections (Algorithm \ref{alg:general_unsup_learning}) is a general algorithm for unsupervised learning from video to detect objects in single images. It presents a sequence of high level steps followed by different modules for an unsupervised learning system. The modules are complementary to each other and function in tandem, each focusing on a specific aspect of the unsupervised learning process. Thus, we have a module for generating data, where soft-masks are produced. There is a module that selects good quality masks.
Then, we have a module for training the next generation classifiers. While, our concept is first presented in high level terms, we also present a specific implementation that represents the first two iterations of the algorithm.
While our implementation is costly during training, in terms of storage and computation time, at test time it is very fast - 0.02 sec per student net
and 0.15 sec per student ensemble.\\

\noindent \textbf{Computation and storage costs.}
During training, the computation time for passing through the teacher pathway during the first iteration of Algorithm \ref{alg:general_unsup_learning} is about 2-3 days: it requires processing data from VID and YTO datasets, including running the VideoPCA module. Afterwards, training the first iteration students, with access to 6 GPUs, takes about 5 days - 6 GPUs are needed for training the 5 different student architectures, since training FConv-Net requires two GPUs in parallel.
Next, training the EvalSeg-Net requires 4 additional days on one GPU. At the second iteration, processing the data through the teacher pathway takes about 3 weeks on 6 GPUs in parallel - it is more costly 
due to the larger training set from which only a small percent (about 10 percent) is selected with EvalSeg-Net. Finally, training the second generation students takes 2 additional weeks. In conclusion, the total computation time required for training, with full access to 6 GPUs is about 7 weeks, when everything is optimized. The total storage cost is about 4TB. At test time the student nets are fast, taking 0.02 sec per image, while the ensemble nets take around 0.15 sec per image.

\section{Experimental analysis}
\label{sec:experiments}

In the first set of experiments we evaluate the impact of the different components of our system.
We experimentally verify that at each iteration the students perform better than their teachers. Then we test the ability of the system to improve from one generation to the next. We also test the effects of data selection and increasing training data size. Then, we compare the performances of each individual network and their combined ensembles.

In Section ~\ref{sec:comparisons}, we compare our algorithm to state of the art methods on object discovery in videos and images. We perform tests on three datasets: YouTube Objects (\cite{prest2012learning}), Object Detection in Internet images (\cite{rubinstein2013unsupervised}) and Pascal-S (\cite{li2014secrets}). In Section ~\ref{sec:transfer_learning} we verify that our unsupervised deep features are also useful in different transfer learning tasks. \\

\noindent \textbf{Datasets.} Unsupervised learning requires large quantities of unlabeled video data. We have chosen for training data, videos from three large datasets: ImageNet VID dataset (\cite{russakovsky2015imagenet}), YouTube Objects (\cite{prest2012learning}) and YouTube Bounding Boxes (\cite{real2017youtube}). VID is one of the largest video datasets publicly available, being fully annotated with ground truth bounding boxes. The dataset consists of about 4000 videos, having a total of
about 1.2M frames. The videos contain objects that
belong to 30 different classes. Each frame could have zero, one or multiple objects annotated. The benchmark challenge associated with this dataset
focuses on the supervised object detection and recognition problem,
which is different from the one that we tackle here. Our system is not trained to identify different object categories, so we do not report results compared to the state of the art on object class recognition and detection, on this dataset. 

YouTube Objects (YTO) is a challenging video dataset with objects undergoing significant changes in appearance, scale and shape, going in and out of occlusion against a varying, often cluttered background. YTO is at its second version now and consists of about 2500 videos, having a total of about 700K frames. 
It is specifically created for unsupervised object discovery, so we perform comparisons to state of the art on this dataset. 

For unsupervised training of our system we used approximately 190k frames from videos chosen from each dataset (120k from VID and 70k from YTO), at learning iteration 1 - those frames which survived after the data selection module. At the second learning iteration, besides improving the classifier, it is important to have access to larger quantities of new unlabeled data. Therefore, for training the second generation of classifiers we added to the unlabeled training set additional 1 million soft-masks, as follows: 600k frames from VID and 400k from the YouTube Bounding Boxes dataset - again, those frames which survived after filtering with the EvalSeg-Net data selection module. Before data selection videos were randomly chosen from each set, VID or YouTube Bounding Boxes, until the total of 1M was reached. We did not add more frames due to heavy computation and storage limitations.\\

\noindent \textbf{Evaluation metrics.} We use different kinds of metrics in our experiments, which depend on the specific task that requires either bounding box fitting or fine segmentation:

\begin{itemize} 

\item \textit{CorLoc} - for evaluating the detection of bounding boxes the most commonly used metric is CorLoc. It is defined
as the percentage of images correctly localized according to the PASCAL criterion:$\frac{B_p \cap B_{GT}}{B_p \cup B_{GT}} \geq 0.5$, where $B_P$ is the predicted bounding box and $B_{GT}$ is the ground truth bounding box. 

\item F-$\beta = \frac{(1-\beta^{2}) precision \times recall}{\beta^{2} \times precision + recall}$ for evaluating the segmentation score on Pascal-S dataset. We use the official evaluation code when reporting results. As in all previous works, we set $\beta^2=0.3$.

\item \textit{P-J metric} P refers to the precision per pixel, while J is the Jaccard similarity (the intersection over union between the output mask the and ground truth segmentations). We use this metric only on Object Discovery in Internet images. For computing the reported results we use the official evaluation code.

\item MAE - Mean Absolute Error is defined as the average pixel-wise difference between the predicted mask and the ground truth. Different from the other metrics, for this metric a lower value is better.

\item \textit{mean IoU} score is defined as $\frac{|G \cap Y|}{|G \cup Y|}$ where $G$ represents the ground truth and $Y$ the predicted mask.

\end{itemize}


\subsection{Evaluation of different system components}


\begin{figure}[ht]
\begin{center}
   \includegraphics[width=1.0\linewidth]{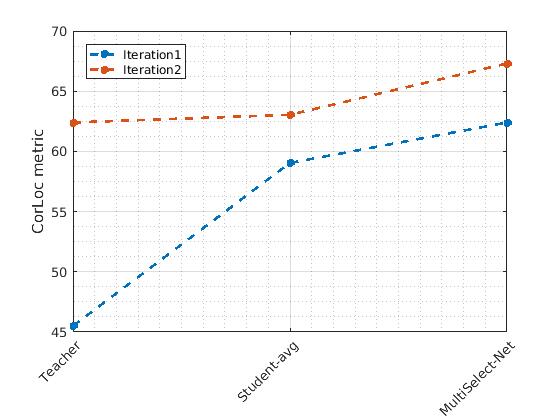}
\end{center}
   \caption{Comparison between the teacher, individual student nets and the ensembles, across two generations (blue line - first iteration; red line - second iteration). Individual students (for which we report average values) outperform the teacher on both iterations, while the ensembles are even stronger than the individual nets. For the second iteration teacher we report the MultiSelect-Net version of the ensemble (since we consider this to be an upper bound). The plots are computed over results on the YouTube Objects dataset using the CorLoc metric (percentage).}
   \label{fig:comparison-teacher}
\end{figure}

\begin{figure}[ht]
\begin{center}
   \includegraphics[width=1.0\linewidth]{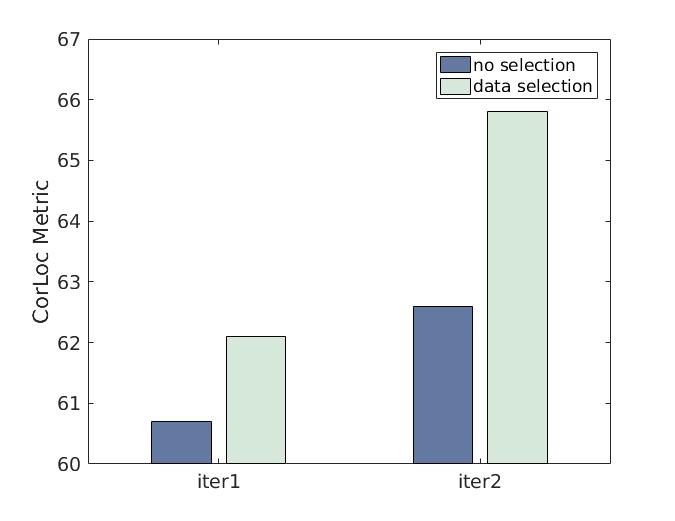}
\end{center}
   \caption{Impact of data selection for both iterations. Data selection (module C) strongly affects the results at each iteration. Note that results from iteration 2 with no selection are slightly better than the ones from iteration 1 with selection. This happens
   because the unlabeled training data is increased and the second generation (iteration 2) teacher pathway is superior, providing better quality masks for training. The results represent the average over 10 classes on YouTube Objects using CorLoc percentage metric.
   }
   \label{fig:influence-of-data-selection}
\end{figure}



\begin{table*}[ht]
\begin{center}
\begin{tabular}{|l|*{6}{c|}|c|c|c|}
\hhline{-------||---}
 & LowRes-Net & FConv-Net & DenseU-Net & BasicU-Net & DilateU-Net  & Avg & Multi-Net & MultiSelect-Net & Avg \\
\hhline{-------||---}
Iteration 1 & 62.1 & 57.6 & 54.6 & 59.1 & 61.8 & 59.0 & 65.3 & 62.4 & 63.9 \\
Iteration 2 & 63.5 & 61.3 & 59.4 & 65.2 & 65.8 & 63.0 & 67.0 & 67.3 & 67.2 \\
\hhline{-------||---}
\noalign{\vspace{\doublerulesep}}
\hhline{-------||---}
Gain &  \color{blue}1.4 \rotatebox{90}{\MVRightarrow}  & \color{blue}3.7 \rotatebox{90}{\MVRightarrow} & \color{blue}4.8 \rotatebox{90}{\MVRightarrow} & \color{blue}6.1 \rotatebox{90}{\MVRightarrow} & \color{blue}4.0 \rotatebox{90}{\MVRightarrow} & \color{blue}4.0 \rotatebox{90}{\MVRightarrow} & \color{blue}1.7 \rotatebox{90}{\MVRightarrow} & \color{blue}4.9 \rotatebox{90}{\MVRightarrow} & \color{blue}3.3 \rotatebox{90}{\MVRightarrow}\\
\hhline{-------||---}
\end{tabular}
\end{center}
\caption{Results of our networks and ensembles on YouTube Objects v1 (\cite{prest2012learning}) dataset (CorLoc metric) at both iterations (generations). We present the average of CorLoc metric of all 10 classes from YTO dataset for each model and ensemble, as well as the average of all single models and the average of the ensembles. As it can be seen, at the second generation there is a clear increase in performance for all models. Also note that at the second generation a single model is able to outperform all the methods (single or ensemble) from the first generation.}
\label{tab:yto_ours}
\end{table*}

\begin{table*}[ht]
\begin{center}
\begin{tabular}{|l|*{6}{c|}|c|c|c|}
\hhline{-------||---}
& LowRes-Net & FConv-Net & DenseU-Net & BasicU-Net & DilateU-Net  & Avg & Multi-Net & MultiSelect-Net & Avg \\
\hhline{-------||---}
Iteration 1 & 85.8 & 79.8 & 83.3 & 86.8 & 85.6 & 84.3 & 85.8 & 86.7 & 86.3 \\
Iteration 2 & 86.7 & 85.6 & 86.7 & 87.1 & 87.9 & 86.8 & 86.4 & 88.2 & 87.3 \\
\hhline{-------||---}
\noalign{\vspace{\doublerulesep}}
\hhline{-------||---}
Gain &  \color{blue}0.9 \rotatebox{90}{\MVRightarrow}  & \color{blue}5.8 \rotatebox{90}{\MVRightarrow} & \color{blue}3.4 \rotatebox{90}{\MVRightarrow} & \color{blue}0.3 \rotatebox{90}{\MVRightarrow} & \color{blue}2.3 \rotatebox{90}{\MVRightarrow} & \color{blue}2.5 \rotatebox{90}{\MVRightarrow} & \color{blue}0.6 \rotatebox{90}{\MVRightarrow} & \color{blue}1.5 \rotatebox{90}{\MVRightarrow} & \color{blue}1.0 \rotatebox{90}{\MVRightarrow}\\
\hhline{-------||---}
\end{tabular}
\end{center}
\caption{Results of our networks and ensemble on Object Discovery in Internet Images (\cite{rubinstein2013unsupervised}) dataset (CorLoc metric) at both iterations (generations). We presented the average CorLoc metric of single models and ensembles per class as well as overall. Note that at the second generation there is a clear increase in performance for all methods. On average, single models from the second iteration are superior to the ensembles from the first.}
\label{tab:internet_ours}
\end{table*}

\begin{table*}[ht]
\begin{center}
\begin{tabular}{|l|*{6}{c|}|c|c|c|}
\hhline{-------||---}
& LowRes-Net & FConv-Net & DenseU-Net & BasicU-Net & DilateU-Net  & Avg & Multi-Net & MultiSelect-Net & Avg\\
\hhline{-------||---}
Iteration 1 & 64.6 & 51.5 & 65.2 & 65.4 & 65.8 & 62.5 & 67.8 & 67.1 & 67.5 \\
Iteration 2 & 66.9 & 61.7 & 68.4 & 68.0 & 67.5 & 66.5 & 69.1 & 68.5 & 68.8 \\
\hhline{-------||---}
\noalign{\vspace{\doublerulesep}}
\hhline{-------||---}
Gain &  \color{blue}2.3 \rotatebox{90}{\MVRightarrow}  & \color{blue}10.2 \rotatebox{90}{\MVRightarrow} & \color{blue}3.2 \rotatebox{90}{\MVRightarrow} & \color{blue}2.6 \rotatebox{90}{\MVRightarrow} & \color{blue}1.7 \rotatebox{90}{\MVRightarrow} & \color{blue}4.0 \rotatebox{90}{\MVRightarrow} & \color{blue}1.3 \rotatebox{90}{\MVRightarrow} & \color{blue}1.4 \rotatebox{90}{\MVRightarrow} & \color{blue}1.3 \rotatebox{90}{\MVRightarrow}\\
\hhline{-------||---}
\end{tabular}
\end{center}
\caption{Results of our networks and ensemble on Pascal-S (\cite{li2014secrets}) dataset (F-$\beta$ metric), for all of our methods for first and second generations as well as their average performance. Note that in this case, since we evaluate actual segmentations and not bounding box fitting, nets with higher resolution output perform better (DenseU-Net, BasicU-Net and DilateU-Net). Again, ensembles outperform single models and the second iteration brings a clear gain in every case.}
\label{tab:pascal_ours}
\end{table*}

\begin{figure*}[ht]
\begin{center}
   \includegraphics[width=1.0\linewidth]{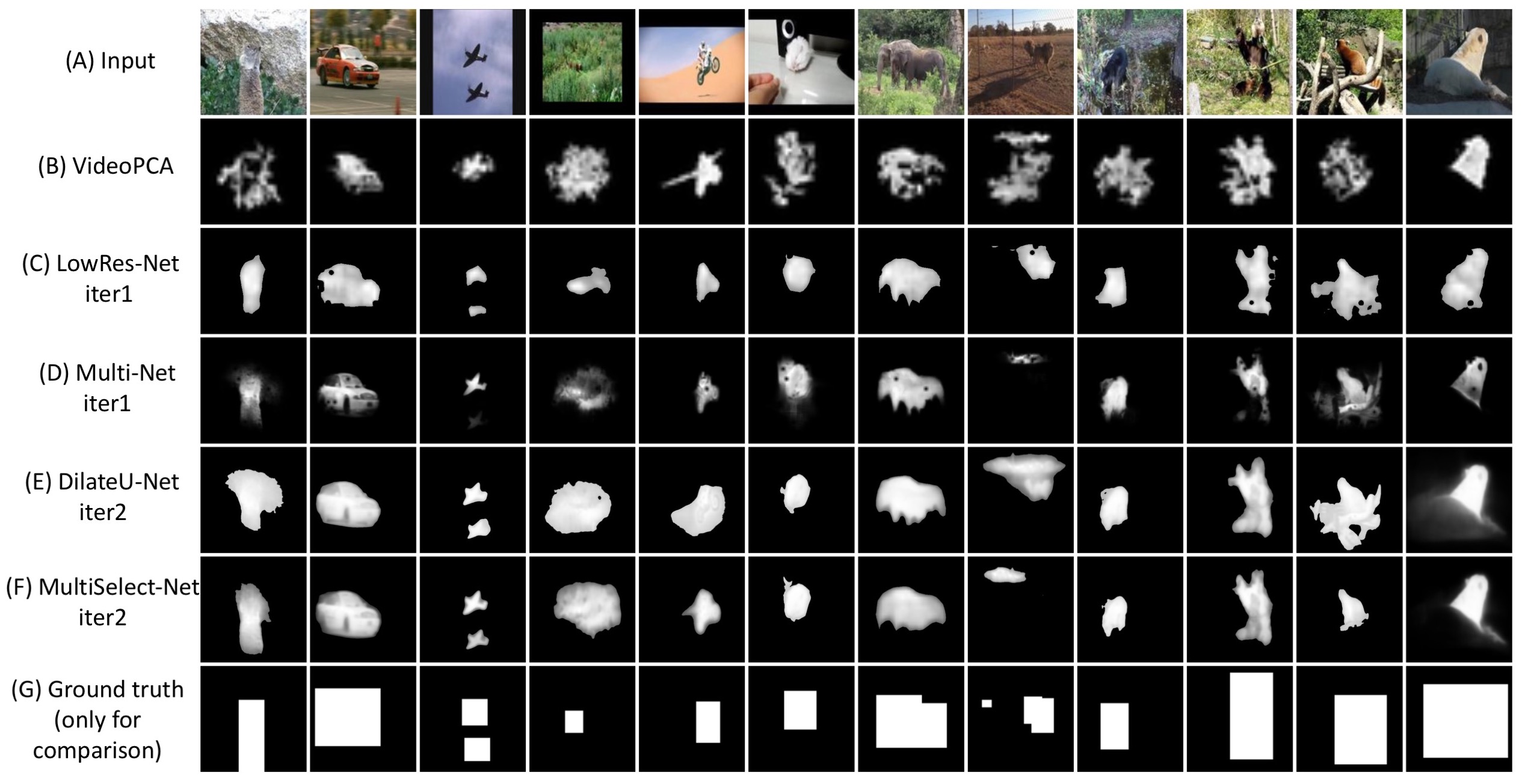}
\end{center}
   \caption{Qualitative results on the VID dataset (\cite{russakovsky2015imagenet}). For each iteration we show results of the best individual and ensemble models, in terms of CorLoc metric. Note the superior quality of our models compared to the VideoPCA (iteration 1 teacher). We also present the ground truth bounding boxes. For more qualitative results please visit our project page \url{https://sites.google.com/view/unsupervisedlearningfromvideo} }
\label{fig:vid_visual}
\end{figure*}

\noindent \textbf{Student vs. teacher} 
In Figure \ref{fig:vid_visual} we present qualitative results on VID dataset as compared to VideoPCA. We can see that the masks produced by VideoPCA are of lower quality, often having holes, non-smooth boundaries and strange shapes. In contrast, the students learn more general shape and appearance characteristics of objects in images, reminding of the grouping principles governing the basis of visual perception as studied by the Gestalt psychologists (\cite{rock1990gestalt}) and the more recent work on the concept of "objectness" (\cite{alexe2010object}). 
The object masks produced by the students are simpler, with very few holes, have nicer and smoother shapes 
and capture well the foreground-background contrast and organization. Another interesting observation is that the students are able to detect multiple objects, a feature that is less commonly achieved by the teacher.

In Figure~\ref{fig:comparison-teacher} we see comparative results between the average of individual models, the ensembles formed and the teacher. Note that the teacher at the next generation reported is the MultiSelect-Net ensemble from the first. We observe that the students at both iterations outperform their respective teachers, which is an interesting and positive outcome. It suggests that we can repeat the process over several iterations and continue to improve. It is also encouraging that the individual nets, which see a single image, are able to generalize and detect objects that are discovered by the teacher in sequences of images. \\

\begin{figure*}[ht]
\begin{center}
   \includegraphics[width=1.0\linewidth]{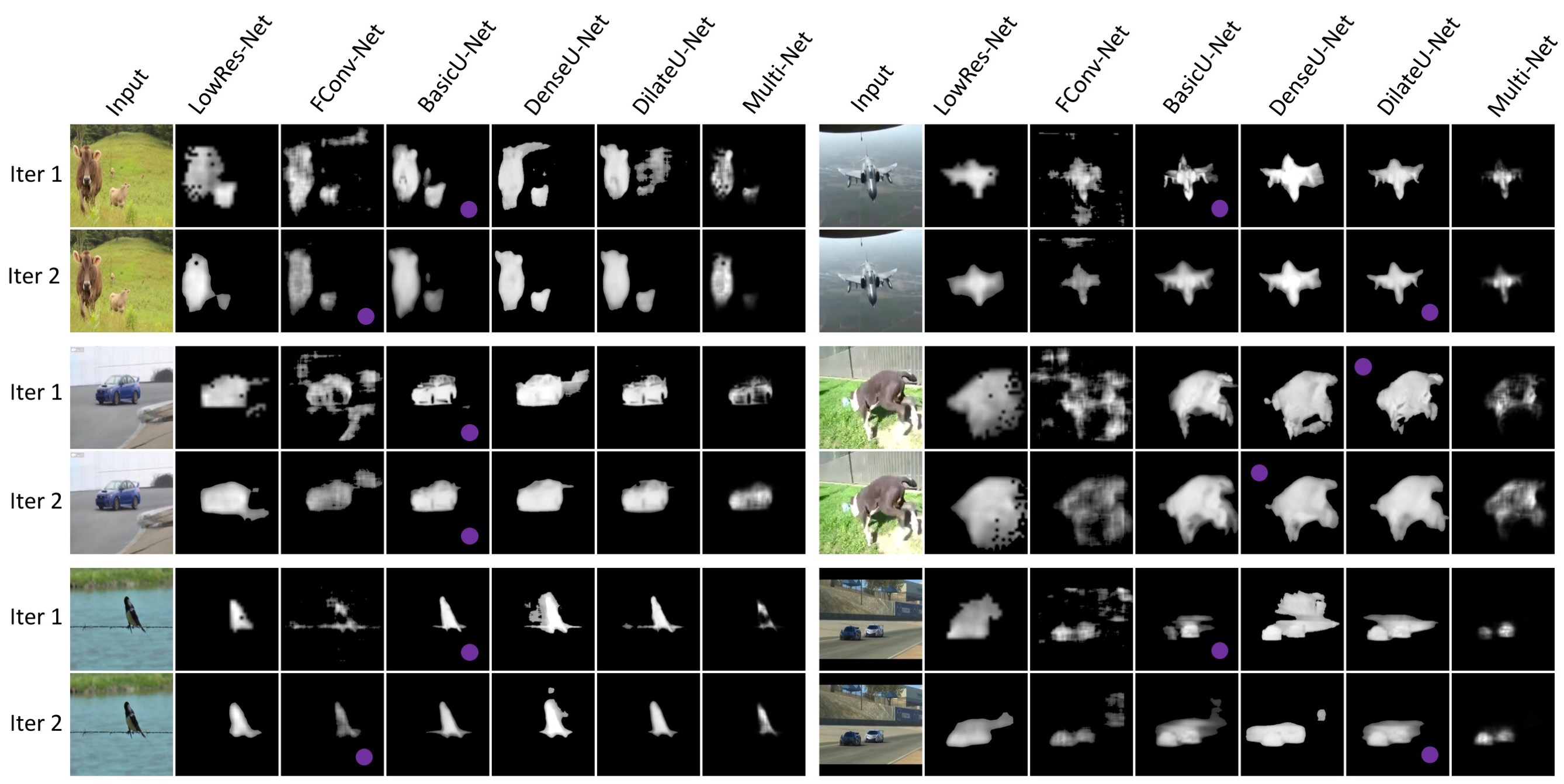}
\end{center}
   \caption{Visual comparison between models at each iteration (generation). We marked with a purple dot the output of our MultiSelect-Net (the top selected student soft mask with EvalSeg-Net). The Multi-Net represents the pixel-wise multiplication between the five models. Note the superior masks at the second generation, with better shapes, fewer holes and sharper edges. }
   \label{fig:comparison-iterations}
\end{figure*}

\noindent \textbf{First vs. next generation.} 
As seen in Tables \ref{tab:yto_ours}, \ref{tab:internet_ours} \ref{tab:pascal_ours} and Figure \ref{fig:comparison-iterations} at the second generation we obtain a clear gain over the first, on all experiments and datasets. This result proves the value of our proposed algorithm that starts from a completely unsupervised object discoverer in video (VideoPCA) and is able to train neural nets for foreground object segmentation, while improving their accuracy over two generations. It uses the students from iteration 1 as teachers at iteration 2. At the second iteration, it also uses more unlabeled training data and it is better at automatically filtering out poor quality segmentations.\\

\noindent \textbf{Impact of data selection.} Data selection is important as seen in Figure~\ref{fig:influence-of-data-selection}. The more selective we are when we accept or reject soft-masks used for training, the better the end result. Also note that being more selective means decreasing the training set. There is a trade-off between selectivity and training data size. \\

\noindent \textbf{Neural architecture vs. data.} 
As seen in Tables \ref{tab:yto_ours}, \ref{tab:internet_ours} and \ref{tab:pascal_ours} different network architecture yield different results, while ensembles always outperform individual models. While the actual CNN architecture has a certain role in performance, another equally important aspect is that of data size. The more data we have the more selective we can afford to be and also the more we could generalize. It is important to increase the data from one generation to the next in order to avoid simply imitating the ensemble of the previous generation. In Tables~\ref{tab:adding_data}  and \ref{tab:adding_data_pj} we show additional tests with our baseline architecture, LowRes-Net, when trained with training sets of different sizes. It is obvious that adding new unlabeled data has a positive effect on performance. The idea of increasing the data in stages is also related to approaches in curriculum learning (\cite{bengio2009curriculum}), where we first learn from easy cases then move to the more complex ones.   \\

\noindent \textbf{Analysis of different ConvNets.}
Our experiments show that different architectures are better at different tasks. LowRes-Net, for example, performs well on the task of box fitting since that does not require a fine sharp object mask. On the other hand, when evaluating the exact segmentation, nets with higher resolution output, which are more specialized for this task perform better. Overall, at the second generation, on box fitting the best single net on average is DilateU-Net and the top ensemble is MultiSelect-Net. However, when it comes to evaluating the actual segmentation the winner is DenseU-Net for single models and Multi-Net for ensembles. In our qualitative results we find that DenseU-Net produces masks with fewer "holes" when compared to DilateU-Net, after thresholding and, thus, it is
better suited for segmentation evaluation. When evaluating the bounding box, these holes do not affect the box and the best model is DilateU-Net. Also, DenseU-Net tends to outputs a mask with higher confidence on the whole object, as opposed to the BasicU-Net and DilateU-Net that output masks with lower confidence around some regions of the object (such as the eyes or wheels). This could be another reason why DenseU-Net produces better segmentations. The model that struggles most during the first iteration is FConv-Net, with significant improvement at the second iteration when the unsupervised training masks are closer to the correct ones.
Also note that the baseline LowRes-Net is a top model on box fitting at the first iteration. The quantitative differences between architectures are shown in Tables \ref{tab:yto_ours}, \ref{tab:internet_ours} and \ref{tab:pascal_ours}, while the qualitative differences can be seen in Figure ~\ref{fig:comparison-iterations}. 


\begin{table}[ht]
\begin{center}
\begin{tabular}{|c|c|c|c|}
\hline
& Training data & CorLoc & Testing dataset \\
\hline
LowRes-Net & VID & 56.1 & \multirow{2}{*}{YTO} \\
LowRes-Net & VID + YTO & 62.2 &  \\
\hline
\end{tabular}
\end{center}
\caption{Influence of adding more unlabeled data, evaluated on YTO with the CorLoc metric. As it can be seen, adding data significantly increases the performance by about 6\%.}
\label{tab:adding_data}
\end{table}

\begin{table}[ht]
\begin{center}
\begin{tabular}{|c|c|c|c|c|}
\hline
& Training data & mean P & mean J \\
\hline
LowRes-Net & VID & 87.73 & 61.25 \\
LowRes-Net & VID + YTO & 88.36 & 62.33 \\
\hline
\end{tabular}
\end{center}
\caption{Influence of adding more unlabeled data on the Object Discovery in Internet images dataset - PJ metric. The performance increases by about 1\%.}
\label{tab:adding_data_pj}
\end{table}


\subsection{Comparisons with state of the art}
\label{sec:comparisons}

\noindent \textbf{Object discovery in video.}
We first performed comparisons with methods specifically designed for object discovery in video. For that, we choose the YouTube Objects dataset and compare it to the best methods on this dataset in the literature (Table \ref{tab:yto}). Evaluations are conducted on both 
versions of YouTube Objects dataset,
YTOv1 (\cite{prest2012learning}) and YTOv2.2 (\cite{kalogeiton2016analysing}). On YTOv1 we follow the same experimental setup as (\cite{jun2016pod, prest2012learning}), by running experiments only on the training videos. We have not included in Table \ref{tab:yto} the results reported by \cite{stretcu2015multiple} because they use a different setup, testing on all videos from YTOv1. It is important to stress out, again, the fact that while the methods presented here for comparison have access to whole video shots, ours only needs a single image at test time. Despite this limitation, our method outperforms the others on 7 out of 10 classes and has the best overall average performance. Note that even our baseline LowRes-Net at the first iteration achieves top performance. The feed-forward CNN processes each image in 0.02 sec, being at least one to two orders of magnitude faster than all other methods (see Table \ref{tab:yto}). We also mention that in all our comparisons, while our system is faster at test time, it takes much longer during its unsupervised training phase and requires large quantities of unsupervised training data. \\

\begin{table*}
\begin{center}
\resizebox{\linewidth}{!}{
\begin{tabular}{|l|*{10}{c|}|c|c|c|}
\hline
Method & Aero & Bird & Boat & Car & Cat & Cow & Dog & Horse & Mbike & Train & Avg & Time & Version \\
\hhline{-----------||---}
\noalign{\vspace{\doublerulesep}}
\hhline{-----------||---}
 \cite{prest2012learning} & 51.7 & 17.5 & 34.4 & 34.7 & 22.3 & 17.9 & 13.5 & 26.7 & 41.2 & 25.0 & 28.5 & N/A &  \\
 \cite{papazoglou2013fast} & 65.4 & 67.3 & 38.9 & 65.2 & 46.3 & 40.2 & 65.3 & 48.4 & 39.0 & 25.0 & 50.1 & 4s & \multirow{6}{*}{v1 }\\
 \cite{jun2016pod} & 64.3 & 63.2 & 73.3 & 68.9 & 44.4 & 62.5 & 71.4 & 52.3 & \textbf{78.6} & 23.1 & 60.2 & N/A & \\
 
 \cite{Haller_2017_ICCV} & 76.3 & 71.4 & 65.0 & 58.9 & \textbf{68.0} & 55.9 & 70.6 & 33.3 & 69.7 & 42.4 & 61.1 & 0.35s & \\
 
\hhline{-----------||--}

LowRes-Net\textsubscript{iter1} & 77.0 & 67.5 & 77.2 & 68.4 & 54.5 & 68.3 & 72.0 & \textbf{56.7} & 44.1 & 34.9 & 62.1 & 0.02s & \\

LowRes-Net\textsubscript{iter2} & 79.7 & 67.5 & 68.3 &	\textbf{69.6} & 59.4 & 75 & \textbf{78.7} & 48.3 & 48.5 & 39.5 & 63.5 & 0.02s & \\

DilateU-Net\textsubscript{iter2} & \textbf{85.1} & \textbf{72.7} & 76.2 & 68.4 & 59.4 & 76.7 & 77.3 & 46.7 & 48.5 & \textbf{46.5} & 65.8 & 0.02s & \\

MultiSelect-Net\textsubscript{iter2} & 84.7 & \textbf{72.7} & \textbf{78.2} & \textbf{69.6} & 60.4 & \textbf{80.0} & \textbf{78.7} & 51.7 & 50.0 & \textbf{46.5} & \textbf{67.3} & 0.15s &\\

\hhline{-----------||---}
\noalign{\vspace{1.5mm}}
\hhline{-----------||---}
\cite{Haller_2017_ICCV} & 76.3 & \textbf{68.5} & \textbf{54.5} & 50.4 & \textbf{59.8} & 42.4 & 53.5 & 30.0 & \textbf{53.5} & \textbf{60.7} & 54.9 & 0.35s & \multirow{6}{*}{v2.2 } \\

\hhline{-----------||--}

LowRes-Net\textsubscript{iter1} & 75.7 & 56.0 & 52.7 & 57.3 & 46.9 & 57.0 & 48.9 & 44.0 & 27.2 & 56.2 & 52.2 & 0.02s &   \\
LowRes-Net\textsubscript{iter2} & 78.1 & 51.8 & 49.0 & 60.5 & 44.8 & 62.3 & 52.9 & 48.9 & 30.6 & 54.6 &	53.4 & 0.02s &   \\
DilateU-Net\textsubscript{iter2} & 74.9 & 50.7 & 50.7 & 60.9 & 45.7 & 60.1 & 54.4 & 42.9 & 30.6 & 57.8 & 52.9 & 0.02s &  \\
BasicU-Net\textsubscript{iter2} & \textbf{82.2} & 51.8 & 51.5 & 62.0 & 50.9 & 64.8 & 55.5 & 45.7 & 35.3 & 55.9 & 55.6 & 0.02s & \\
MultiSelect-Net\textsubscript{iter2} & 81.7 & 51.5 & 54.1 & \textbf{62.5} & 49.7 & \textbf{68.8} & \textbf{55.9} & \textbf{50.4} & 33.3 & 57.0 & \textbf{56.5} & 0.15s & \\
\hline
\end{tabular}}
\end{center}
\caption{Results on Youtube Objects dataset, versions v1 (\cite{prest2012learning}) and v2.2 (\cite{kalogeiton2016analysing}). We achieve state of the art results on both versions. Please note that the baseline LowRes-Net already achieves top results on v1, while being close to the best on v2.2. We present results of the top individual and ensemble models and also keep the baseline LowRes-Net at both iterations, for reference. Note that complete results on this dataset v1 for all models are also presented in Table \ref{tab:yto_ours}. }
\label{tab:yto}
\end{table*}



\noindent \textbf{Object discovery in images} 
We compare our system against other methods that perform image discovery in images. We use two different datasets for this comparison: Object Discovery in Internet Images and Pascal-S datasets. We report results using metrics that are commonly used for these tasks, as presented at the beginning of the experimental section.

Object Discovery in Internet Images is a representative benchmark for foreground object detection in single images. This set contains internet images and it is annotated with high detail segmentation masks. In order to enable comparison with previous methods, we use the 100 images subsets provided for each of the three categories: airplane, car and horse. The methods evaluated on this dataset in the literature, aim to either discover the bounding box of the main object in a given image or its fine segmentation mask. We evaluate our system on both. Note that different from other works, we do not need a collection of images during test time, since each image can be processed independently by our system. Therefore, unlike other methods, our performance is not affected by the structure of the image collection or the number of classes of interest being present in the collection.

In Table \ref{tab:object_discovery} we present the performance of our method as compared to other unsupervised object discovery methods in terms of CorLoc on the Object Discovery dataset. We compare our predicted box against the tight box fitted around the ground-truth segmentation as done in \cite{cho2015unsupervised,tang2014co}.
Our system can be considered in the mixed class category: it does not depend on the structure of the image collection. It treats each image independently. The performance of the other algorithms degrades as the number of main categories increases in the collection (some are not even tested by their authors on the mixed-class case), which is not the case with our approach.

We obtain state of the art
results on all classes, improving by a significant margin over the method of \cite{cho2015unsupervised}. When the method in \cite{cho2015unsupervised} is allowed to see a collection of images that are limited to a single majority class, its performance improves and it is equal with ours on one class. However, our method has no other information necessary besides the input image, at test time.


\begin{table}[t]
\begin{center}
\begin{tabular}{|l|c|c|c|c|}
\hline
Method & Airplane & Car & Horse & Avg\\
\hline\hline
\cite{kim2011distributed} & 21.95 & 0.00 & 16.13 & 12.69\\
\cite{joulin2010discriminative} & 32.93 & 66.29 & 54.84 & 51.35 \\
\cite{joulin2012multi} & 57.32 & 64.04 & 52.69 & 58.02\\
\cite{rubinstein2013unsupervised} & 74.39 & 87.64 & 63.44 & 75.16\\
\cite{tang2014co} & 71.95 & 93.26 & 64.52 & 76.58\\
\cite{cho2015unsupervised} & 82.93 & 94.38 & \color{red}\textit{75.27} & 84.19\\
\cite{cho2015unsupervised} mixed & 81.71 & 94.38 & 70.97 & 82.35\\
\hline

LowRes-Net\textsubscript{iter1} & 87.80 & \textbf{95.51} & 74.19 & 85.83 \\

LowRes-Net\textsubscript{iter2} & 93.90 & 92.13 & 74.19 & 86.74 \\

DilateU-Net\textsubscript{iter2} & \textbf{95.12} & \textbf{95.51} & 73.12 & 87.92 \\

MultiSelect-Net\textsubscript{iter2} &  93.90 & \textbf{95.51} & \textbf{75.27} & \textbf{88.22} \\

\hline
\end{tabular}
\end{center}
\caption{Results on the Object Discovery in Internet images (\cite{rubinstein2013unsupervised}) dataset (CorLoc metric). The results obtained in the first iteration are further improved in the second one. 
We present the best single and ensemble models, along with the baseline LowRes-Net at both iterations. Among the single models DilateU-Net is often the best when evaluating box fitting.}
\label{tab:object_discovery}
\end{table}

\begin{table}[t]
\begin{center}
\resizebox{\linewidth}{!}{
\begin{tabular}{|c|c|c|c|c|c|c|}
\hline
\multirow{2}{*}{} & \multicolumn{2}{c|}{Airplane} & \multicolumn{2}{c|}{Car} & \multicolumn{2}{c|}{Horse} \\
\cline{2-7}
& P & J & P & J & P & J \\
\hline\hline
\cite{kim2011distributed} & 80.20 & 7.90 & 68.85 & 0.04 & 75.12 & 6.43 \\
\cite{joulin2010discriminative} & 49.25 & 15.36 & 58.70 & 37.15 & 63.84 & 30.16 \\
\cite{joulin2012multi} & 47.48 & 11.72 & 59.20 & 35.15 & 64.22 & 29.53 \\
\cite{rubinstein2013unsupervised} & 88.04 & 55.81 & 85.38 & 64.42 & 82.81 & 51.65 \\
\cite{chen2014enriching} & 90.25 & 40.33 & 87.65 & 64.86 & 86.16 & 33.39 \\

\hline
LowRes-Net\textsubscript{iter1} & \textbf{91.41} & 61.37 & 86.59 & 70.52 & 87.07 & 55.09 \\

LowRes-Net\textsubscript{iter2} & 90.61 & 60.19 & 87.05 & 71.52 & 88.73 & 55.31 \\

DenseU-Net\textsubscript{iter2} & 91.03 & 64.46 & 85.71 & 72.51 & 87.14 & \textbf{55.44} \\

Multi-Net\textsubscript{iter2} & 91.13 & \textbf{66.02} & \textbf{87.67} & \textbf{73.98} & \textbf{88.83} & 55.23 \\

\hline
\end{tabular}}
\end{center}
\caption{Results on the Object Discovery in Internet images (\cite{rubinstein2013unsupervised}) dataset using (P, J metric) on segmentation evaluation. We present results of the top single and ensemble models, along with LowRes-Net at both iterations. On the task of fine object segmentation the best individual model tends to be DenseU-Net as also mentioned in the text. Note that we applied GrabCut on these experiments only as a post-processing step, since all methods reported in this Table also used it.}
\label{tab:object_discovery_pj}
\end{table}

\begin{figure*}[ht]
\begin{center}
   \includegraphics[width=1.0\linewidth]{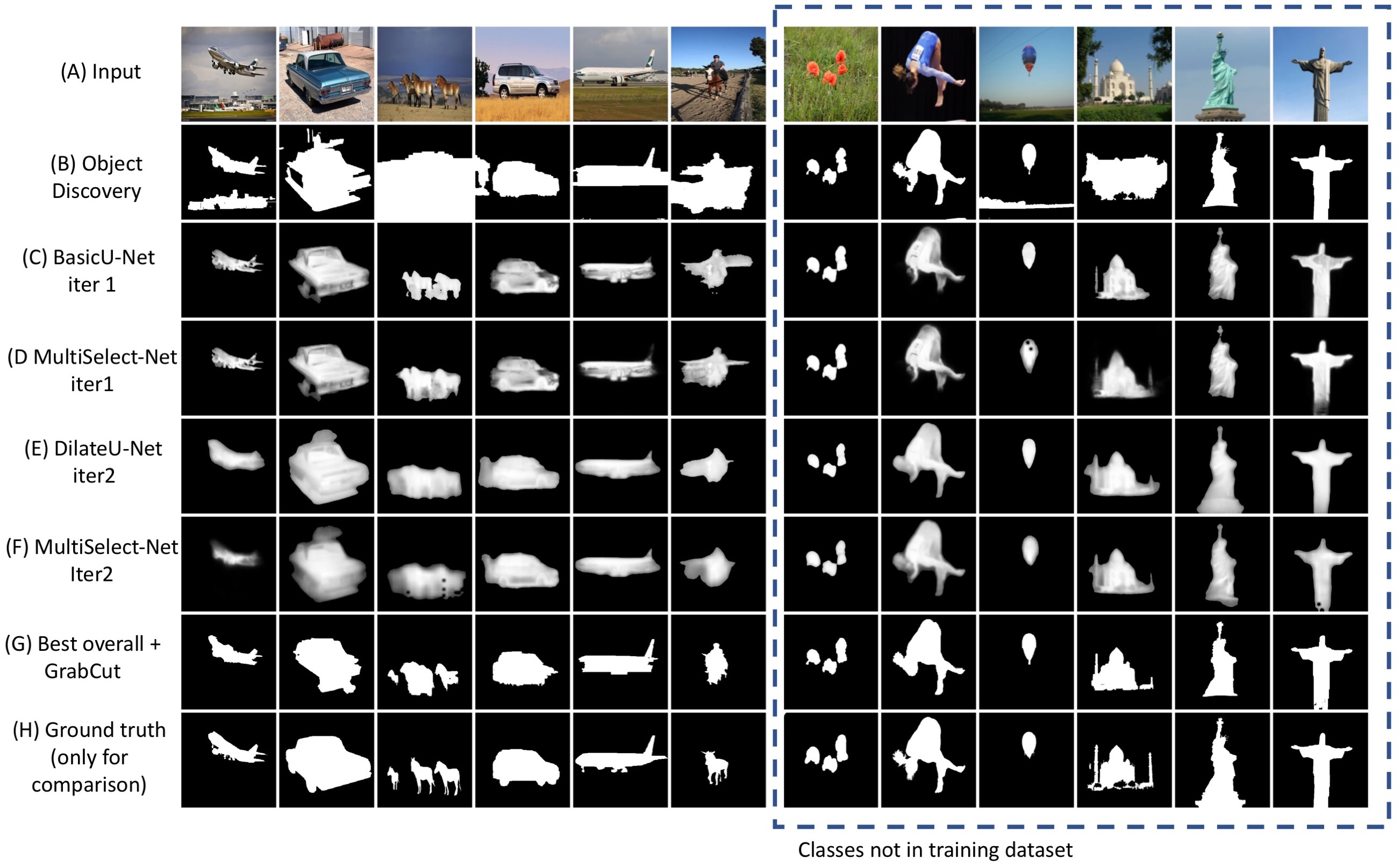}
\end{center}
   \caption{Qualitative results on the Object Discovery dataset as compared to (B)~\cite{rubinstein2013unsupervised}. For both iterations we present results of the top single and ensemble models (C-F), without using GrabCut. We also present results when GrabCut is used with the top ensemble (G). Note that our models are able to segment objects from classes that were not present in the training set (examples on the right side). Also, note that the initial VideoPCA teacher cannot be applied on single images.
  }
\label{fig:internet_visual}
\end{figure*}

\begin{figure*}
\begin{center}
   \includegraphics[width=1.0\linewidth]{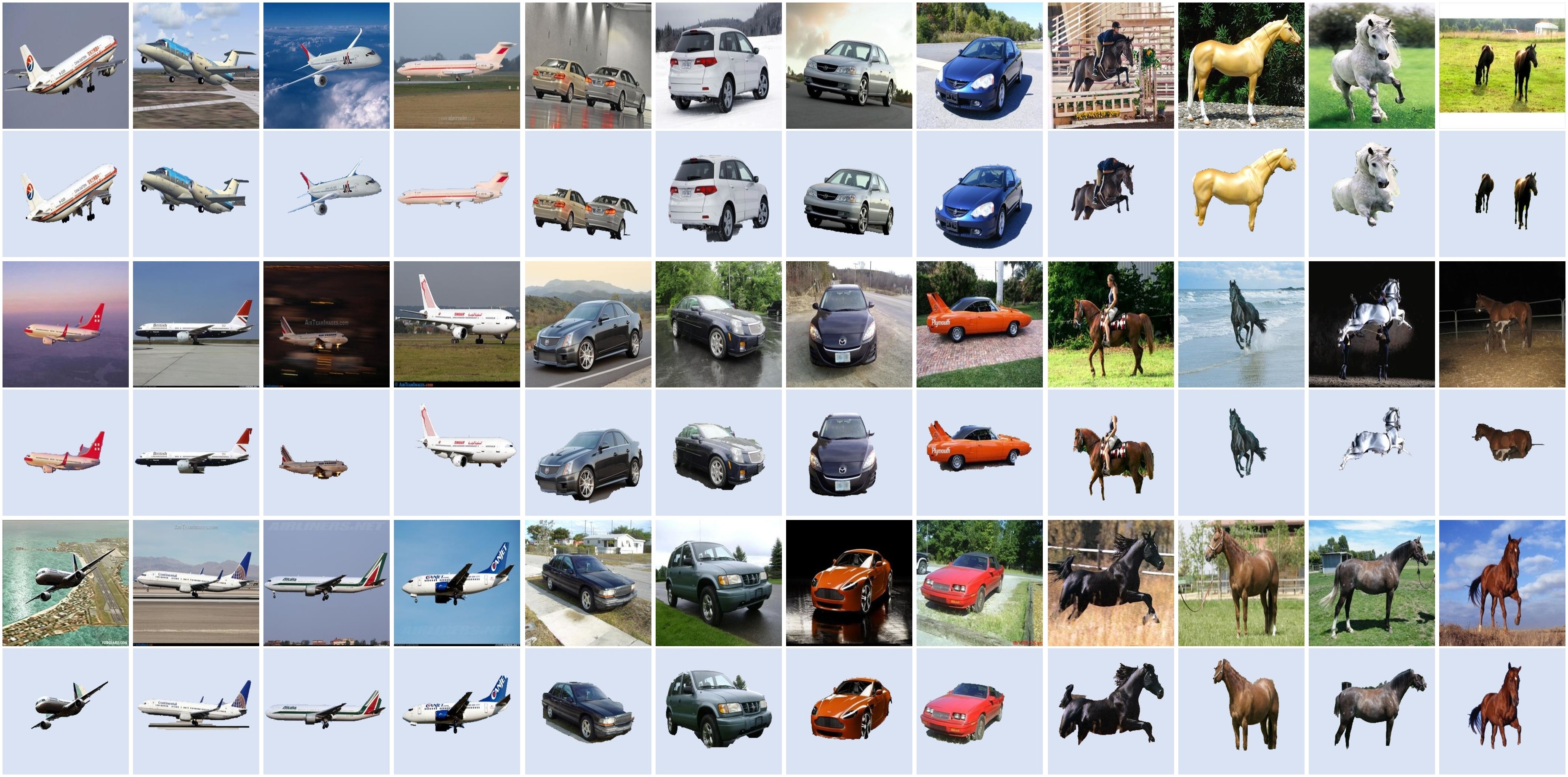}
\end{center}
   \caption{Qualitative results on the Object Discovery in Internet Images \cite{rubinstein2013unsupervised} dataset. For each example we show the input RGB image and immediately below our segmentation result,  with GrabCut post processing for obtaining a hard segmentation. Note that our method produces good quality segmentation results, even in images with cluttered background.}
\label{fig:visual}
\end{figure*}

We also tested our method on the task of fine foreground object segmentation and compared to the best performers in the literature on the Object Discovery dataset in Table \ref{tab:object_discovery_pj}. For refining our soft masks we apply the GrabCut method, as it is available in OpenCV. We evaluate based on the same
P, J evaluation metric as described by  \cite{rubinstein2013unsupervised} - the higher P and J, the better. In Figure \ref{fig:internet_visual} and \ref{fig:visual} we present some qualitative results for each class. As mentioned previously, these experiments on Object Discovery in Internet Images are the only ones on which we apply GrabCut as a post-processing step, as also used by all competing methods presented in
Table \ref{tab:object_discovery_pj}.

Another important dataset used for the evaluation of a related task, that of salient object detection, is Pascal-S dataset, consisting of 850 images. As seen from Table~\ref{tab:pascalS_results} we achieve top results on all three metrics against methods that do not use any supervised pre-trained features. Being a foreground object detection method, our approach is usually biased towards the main object in the image - even though it can also detect multiple ones. Images in Pascal-S usually have more objects, so we consider our results very encouraging being close to approaches that use features pre-trained in a supervised manner. Also note that we did not use GrabCut for these experiments.

On single image experiments, our system was
trained, as discussed before on other, video datasets (VID, YTO and YTB). It has not previously seen any of the images in Pascal-S or Object Discovery datasets during training.

\begin{table}[ht]
\begin{center}
\begin{tabular}{|l|c|c|c|c|l|}
\hline
\multirow{3}{*}{Method} & \multirow{3}{*}{$F_\beta$} & \multirow{3}{*}{MAE} & \multicolumn{1}{|p{1cm}|}{\vspace{2pt} \centering mean \\ IoU} & \multicolumn{1}{|p{2cm}|}{\centering pre-trained \\ supervised features?}\\
\hline\hline
\cite{wei2012geodesic} & 56.2 & 22.6 & 41.6 & no \\ 
\cite{li2015weighted} & 56.8 & 19.2 & 42.4 & no \\ 
\cite{zhu2014saliency} & 60.0 & 19.7 & 43.9 & no \\ 
\cite{yang2013saliency} & 60.7 & 21.7 & 43.8 & no \\ 
\cite{zhang2015minimum} & 60.8 & 20.2 & 44.3 & no \\ 
\cite{tu2016real} & 60.9 & 19.4 & 45.3 & no \\ 
\cite{zhang2017supervision} & 68.0 & \color{red}\textit{14.1} & \color{red}\textit{54.9} & init VGG \\ 
\hline
LowRes-Net\textsubscript{iter1} & 64.6 & 19.6 & 48.7 & no\\
LowRes-Net\textsubscript{iter2} & 66.9 & 18.3 & 51.4 & no\\
DenseU-Net\textsubscript{iter2} & 68.4 & \textbf{17.6} & 51.6 & no \\
Multi-Net\textsubscript{iter2} & \textbf{69.1} & 19.2 & \textbf{53.0} & no \\

\hline
\end{tabular}
\end{center}
\caption{Results on the PASCAL-S dataset compared against other unsupervised methods. For MAE score lower is better, while for $F_\beta$ and mean IoU higher is better. We reported max {$F_\beta$}, min MAE and max mean IoU for every method. In bold we presented the top results when no supervised pre-trained features were used.}
\label{tab:pascalS_results}
\end{table}



\subsection{Transfer learning experiments}
\label{sec:transfer_learning}

While the focus of the paper is foreground object detection in the unsupervised learning setup, we also want to verify the usefulness of our approach on transfer learning experiments. We design experiments to test two aspects of our system - the actual unsupervised features learned and the final output foreground mask. We perform tests on YouTube Objects v1 dataset, in a relatively standard supervised classification setup, by learning to classify individual video frames with the class given by their parent video shot - for a total of ten classes.

We use the frames from the YTO training videos for training and the ones from the YTO test videos for testing. We test on a frame by frame basis and report the average multiclass classification percentage - how often the correct class is chosen out of ten classes.
This problem is difficult for several reasons: 1) the training and testing frames come from different videos in YTO, that vary significantly in 
appearance and background scene 2) the object of interest is not present in every frame, which makes the classification rely heavily on the contextual scene. 3) there are multiple objects in many frames, having a cluttered background, while the object of interest goes through different changes in scale, viewpoint and pose. 

We have two experimental setups for this task, one focused on the pre-trained features and the other on the foreground masks. In the first setup, we replace the last fully connected layer from our baseline model LowRes-Net with a classification part and freeze the network up to a given depth, using as pre-trained features the ones from the unsupervised learning task. Then, we fine-tune the end part on the given supervised classification task. In the second experimental setup we extract features from VGG network pre-trained (\cite{simonyan2014very}) on ImageNet from different subwindows of the image, one being the bounding box given by the unsupervised LowRes-Net. Both tests that are presented next in more detail, prove that our approach is useful on transfer learning tasks.\\

\begin{figure}[t]
\begin{center}
   \includegraphics[width=0.9\linewidth]{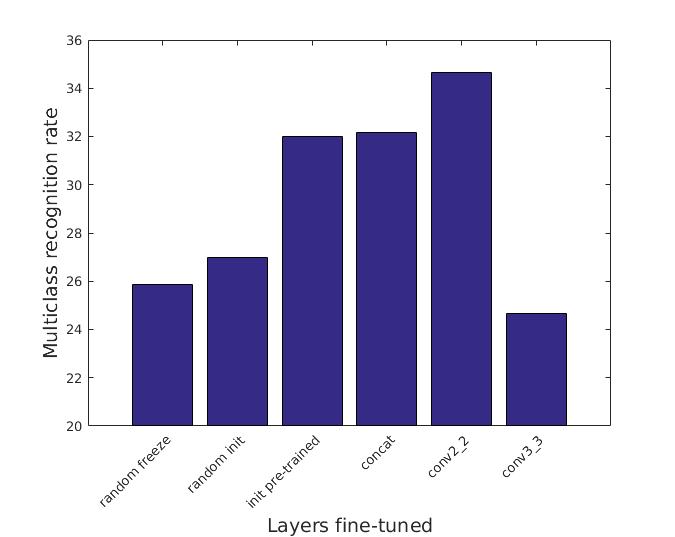}
\end{center}
   \caption{Transfer learning with pre-trained unsupervised features: the "random freeze" case is when the network is randomly initialized then frozen and the classification part is trained; "random init" is when the whole network and the classification part are randomly initialized and then trained jointly, end-to-end; "concat", "conv2\_2" and "conv3\_3" refer to cases where the network trained in the unsupervised way is frozen up to and including that specified layer, and the rest, including the classification part, is then trained; the "init pre-trained" case is when the network is initialized with the pre-trained features from LowRes-Net then everything fine-tuned on the classification task. The results indicate that the optimal case is when unsupervised pre-trained features from the middle part are used, which are more likely to be relevant for different semantic classes than the last deep features that produce the final segmentation.}
\label{fig:transfer_learning}
\end{figure}

\noindent \textbf{Using the unsupervised features.} In this experimental setup, we replace the last fully connected layer with classification part, composed of a 
reduction convolutional layer having four filters and a final fully connected layer with 10 neurons. We test various cases by freezing different parts of the LowRes network and fine-tune the rest on the supervised classification task. The results are presented in Figure \ref{fig:transfer_learning}. 

They strongly suggest that the features learned in an unsupervised way from the middle of the network are best suited for semantic classification. The result clearly demonstrates the usefulness of the unsupervised features on the supervised classification task. In all cases when these features are used the results are improved ("concat",  "conv2\_2", "init pre-trained") except for one case, "conv3\_3". This happens because the pretrained features used in this case are from the top level - when the final segmentation is produced. At that level the semantic information is already lost. On the contrary, when features are frozen at the middle of the network, the best results are obtained.
\\

\noindent \textbf{Using the detected foreground bounding box.}
In these experiments we extract 'fc7' VGG19 features, pre-trained on ImageNet, by passing through VGG19 different subwindows of the image rescaled appropriately, namely the whole image, the center box with height and width being half the original image size and the window cropped according to the bounding box produced by LowRes-Net. We concatenate such features taken from these windows in different combinations and pass them through a last fully connected layer with 10 neurons, which we train on the given classification task. We then, test the different combinations as shown in Table \ref{tab:guided_learning}. 
When using features extracted from the bounding-box fitted with LowRes-Net (alone or in combination with the whole image), we obtain significantly better results compared to the case when windows are extracted from fixed locations only (middle box, whole image or in combination). These results verify that the foreground segmentation mask detected with our models is, as expected, directly related to the main video class and constitutes a valuable source of information in image classification tasks.

Overall, the classification experiments presented in this Section indicate that the features learned in an unsupervised manner with our algorithm contain relevant semantic information about object classes and could be useful for related supervised learning tasks.

\begin{table}[t]
\begin{center}
\begin{tabular}{|l|c|}
\hline
Region of extracted features & Multiclass recognition rate \\
\hline\hline

Whole image & 69.1 \\
Middle crop image & 64.9 \\
Cropped image by LowRes-Net & 70.2 \\
Whole + middle crop & 67.2 \\
Whole + cropped by LowRes-Net & \textbf{72.7} \\

\hline
\end{tabular}
\end{center}
\caption{Classification experiments using the foreground mask.
Different sub-windows of the image are passed through VGG19 and features are extracted for the given classification task. Note that a significant boost is obtained when features are also extracted from the bounding box fitting based on the soft-mask predicted by LowRes-Net.}
\label{tab:guided_learning}
\end{table}

\section{Short discussion on unsupervised learning}

The ultimate goal of unsupervised learning might not be about matching the performance of the supervised case but rather about reaching beyond the capabilities of the classical supervised scenario. An unsupervised system should be able to learn and recognize different object classes, such as animals, plants and man-made objects, as they evolve and change over time, from the past and into the unknown future. It should also be able to learn about new classes that might be formed, in relation to others, maybe known ones. We see this case as fundamentally different from the supervised one in which the classifier is forced to learn from a distribution of samples that is fixed and limited to a specific period of time - that when the human labeling was performed.

Therefore, in the supervised learning paradigm a car from the future, should not be classified as car, because it is not a car, according to the supervised distribution of cars given at present training time, when human annotations are collected. On the other hand, a system that learns by itself should be able to track how cars have been changing in time and recognize such objects as "cars" - with no step by step human intervention.

From a temporal perspective, unsupervised learning is about continuous learning and adaptation to huge quantities of data that are perpetually changing. Human annotation is extremely limited in an ocean of data and not able to provide the so called "ground truth" information continuously. Therefore, unsupervised learning will soon become a core part, larger than the supervised one, in the future of artificial intelligence.

\section{Conclusions and future work}
\label{sec:conclusions}

In this article, we present a novel and effective approach to learning from video, in an unsupervised fashion, to detect foreground objects in single images. We present a relatively general algorithm for this task, which offers the possibility of learning several generations of students and teachers. We demonstrate in practice that the system improves its performance over the course of two generations. We also test the impact of the different system components on performance and show state of the art results on three different datasets. To our best knowledge, it is the first system that learns to detect and segment foreground objects in images in an unsupervised fashion, with no pre-trained features given or manual labeling, while requiring only a single image at test time. 

The convolutional networks trained along the student pathway are able to learn general "objectness" characteristics, which include good form, closure, smooth contours, as well as contrast with the background. What the simpler initial VideoPCA teacher discovers over time, the deep, complex student is able to learn across several layers of image features at different levels of abstraction. Our results on transfer learning experiments are also encouraging and show additional cases in which such a system could be useful. In future work we plan to further grow our computational and storage capabilities to demonstrate the power of our unsupervised learning algorithm along many generations of student and teacher networks.
We believe that our approach, tested here in extensive experiments, will bring a valuable contribution to computer vision research.

\begin{acknowledgements}
This work was supported by UEFISCDI, under projects PN-III-P4-ID-ERC-2016-0007, PN-III-P2-2.1-PED-2016-1842 and PN-III-P1-1.2-PCCDI-2017-0734.
\end{acknowledgements}

\bibliographystyle{spbasic}      
\bibliography{bib_ijcv17_CBL}

\end{document}